%% file: OneSign.tex
\documentclass{article}

\usepackage{microtype}
\usepackage{graphicx}
\usepackage{subfigure}
\usepackage{booktabs} 

\usepackage{hyperref}
\usepackage{multirow}
\usepackage{CJKutf8}

\usepackage[preprint]{icml2026}

\usepackage{amsmath}
\usepackage{amssymb}
\usepackage{mathtools}
\usepackage{amsthm}
\usepackage[utf8]{inputenc} 

\usepackage[capitalize,noabbrev]{cleveref}
 
\theoremstyle{plain}

\theoremstyle{definition}

\theoremstyle{remark}

\usepackage{caption}
\usepackage[table]{xcolor}
\usepackage{colortbl}
\usepackage{float}
\usepackage{wrapfig}

\usepackage[textsize=tiny]{todonotes}

\input{preamble}

\icmltitlerunning{OneSign: Unifying Sign Language Understanding Tasks with One Model}

\begin{document}

\twocolumn[
\icmltitle{OneSign: Unifying Sign Language Understanding Tasks with One Model}



\begin{icmlauthorlist}

\icmlauthor{Shiwei Gan}{nju}
\icmlauthor{Yafeng Yin}{nju}
\icmlauthor{Xiao Liu}{nju}
\icmlauthor{Desibieer Tuerdaken}{nju} 
\icmlauthor{Lei Xie}{nju}
\icmlauthor{Sanglu Lu}{nju}

\end{icmlauthorlist}

\icmlaffiliation{nju}{
State Key Laboratory of Novel Software Technology,
Nanjing University,
Nanjing 210023,
China
}

\icmlcorrespondingauthor{Yafeng Yin}{yafeng@nju.edu.cn}

\icmlkeywords{Sign Language}

\vskip 0.3in
]



\printAffiliationsAndNotice{}  

\begin{abstract}
\input{Section/abstract}

\end{abstract}

\section{Introduction} 
\label{sec:introduction}

\input{Section/introduction}

\vspace{-2mm}
\section{Related Work}
\label{sec:related}
\input{Section/related}

\section{Method}
\label{sec:Method}

\input{Section/method}

\input{Section/performance}

\section{Conclusion}
\label{sec:conclusion}
\input{Section/conclusion}

 \bibliography{bib/slt,bib/cslr,bib/islr,bib/other,bib/dataset}
\bibliographystyle{icml2026}

\newpage
\appendix
\onecolumn

\input{Section/sup}

\end{document}

%% file: preamble.tex
\usepackage{graphicx} 
\usepackage{booktabs}
\usepackage{multibib}
\usepackage{CJKutf8}
\usepackage{balance}
\usepackage{caption}
\usepackage{subcaption}
\usepackage{multirow}  
\usepackage{amsthm,amsmath,amssymb}

\usepackage{mathrsfs}
\usepackage{bbding}
\usepackage{url}            %
\usepackage[table]{xcolor}

\newlength\savewidth\newcommand\shline{\noalign{\global\savewidth\arrayrulewidth \global\arrayrulewidth 1pt}\hline\noalign{\global\arrayrulewidth\savewidth}}
\newcommand{\tablestyle}[2]{\setlength{\tabcolsep}{#1}\renewcommand{\arraystretch}{#2}\centering\footnotesize}

\makeatletter\renewcommand\paragraph{\@startsection{paragraph}{4}{\z@}
  {.5em \@plus1ex \@minus.2ex}{-.5em}{\normalfont\normalsize\bfseries}}\makeatother

\newcolumntype{x}[1]{>{\centering\arraybackslash}p{#1pt}}
\newcolumntype{y}[1]{>{\raggedright\arraybackslash}p{#1pt}}
\newcolumntype{z}[1]{>{\raggedleft\arraybackslash}p{#1pt}}

\newcommand{\app}{\raise.17ex\hbox{$\scriptstyle\sim$}}

\definecolor{baselinecolor}{gray}{.9}
\newcommand{\baseline}[1]{\cellcolor{baselinecolor}{#1}}
 
\newcolumntype{^}{>{\currentrowstyle}}

\definecolor{dt}{gray}{0.7}  %

\usepackage[capitalize]{cleveref}
\crefname{section}{Sec.}{Secs.}
\Crefname{section}{Section}{Sections}
\Crefname{table}{Table}{Tables}
\crefname{table}{Tab.}{Tabs.}

\newcolumntype{S}{@{}>{\lrbox0}l<{\endlrbox}}  %
\definecolor{lightgreen}{HTML}{D8ECD1}

\newcommand{\ie}{{\emph{i.e.}}, }

\newcommand{\eg}{{\emph{e.g.}}, }

\usepackage{tikz}

\definecolor{deemph}{gray}{0.6}
\newcommand{\gc}[1]{\textcolor{deemph}{#1}}

\usepackage{CJKutf8}

%% file: Section/abstract.tex
Sign Language Understanding (SLU) encompasses a diverse set of tasks, including Isolated Sign Language Recognition (ISLR), Continuous Sign Language Recognition (CSLR), and Sign Language Translation (SLT). Although these tasks share basic semantic and linguistic foundations, they are typically addressed with task-specific architectures and training pipelines, which hinders knowledge sharing and requires costly pretraining and finetuning for each task.
In this paper, we focus on two aspects of SLU tasks: 
(1) training and inference pipelines are highly fragmented: most  methods rely on pretraining on large-scale SL datasets followed by task- or dataset-specific finetuning, which leads to multiple specialized models rather than a single  checkpoint. (2) current LLM-based methods  may overlook the inherent modality discrepancy between sign and text tokens, simply concatenating them and processing both modalities with the same decoder layers.
In this paper, we present OneSign, a unified framework that addresses multiple SLU tasks within a single model and a single checkpoint. OneSign reformulates ISLR, CSLR, and SLT under a single training paradigm, allowing all tasks to be trained and inferred using the same model checkpoint without dataset or task-specific pretraining or retraining. 
To accommodate the heterogeneous characteristics of sign and text representations, we introduce a \textbf{Modality-Adaptive Mixture-of-Experts}(MA-MoE) architecture, consisting of a shared expert and modality-specific experts for sign and text tokens. A modality router dynamically activates the corresponding experts, and their outputs are aggregated to form the final token representations. By enabling modality-dependent expert specialization while preserving a shared expert path, MA-MoE can effectively model the modality differences between continuous sign representations and discrete text tokens. Extensive experiments on multiple benchmarks demonstrate that OneSign achieves competitive or state-of-the-art performance on several benchmarks, highlighting its effectiveness as a unified SLU model. Datasets are available at : \url{https://github.com/gswycf/OneSign}.

\begin{figure}
    \centering \includegraphics[width=0.85\linewidth]{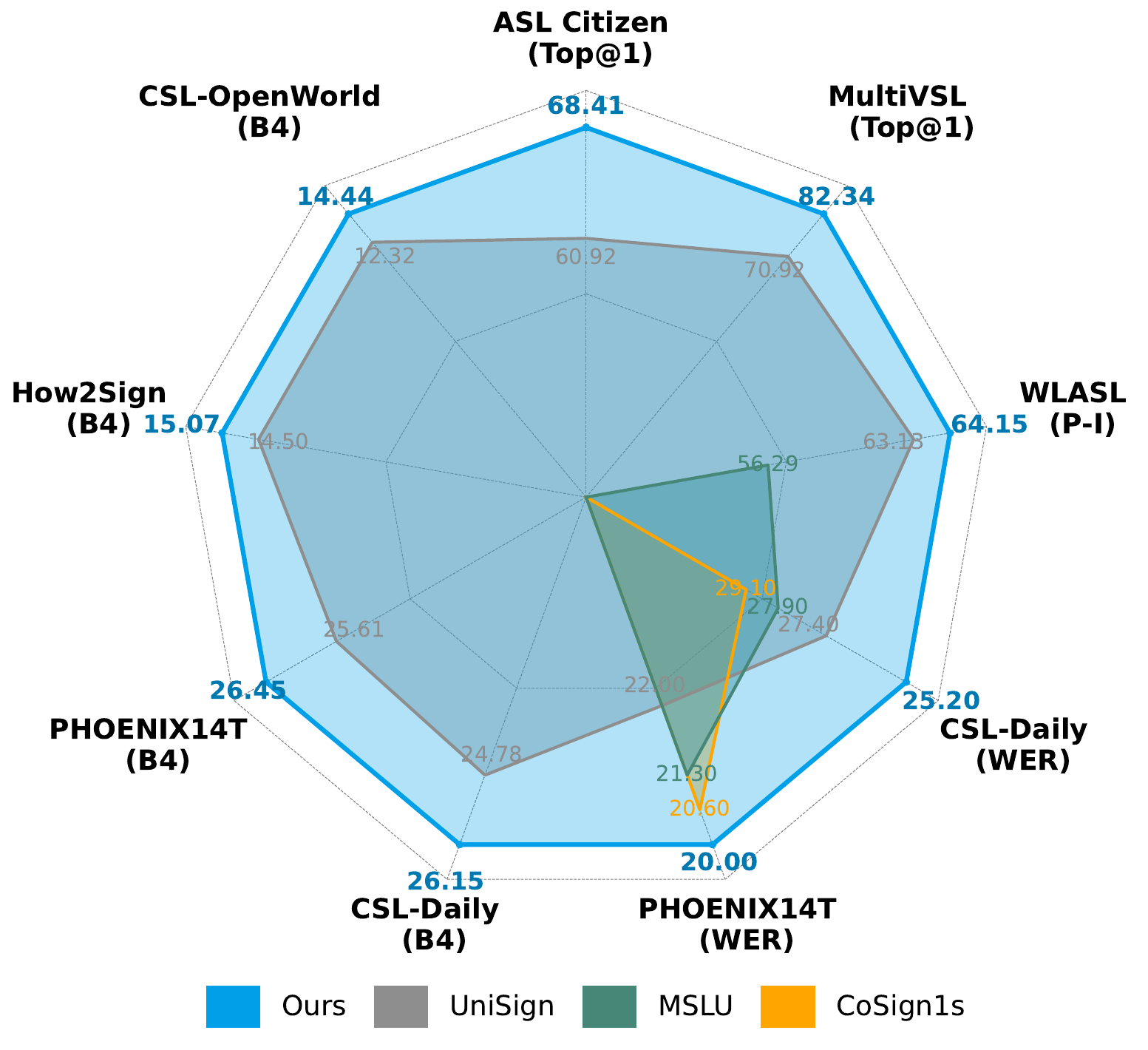}
    \caption{Overall performance comparison across multiple tasks.}
    \label{fig:overall}
    \vspace{-3mm}
\end{figure}

%% file: Section/introduction.tex
Sign Language Understanding (SLU) encompasses a family of closely related tasks~\cite{gan2024signgraph,gan2023towards,gan23contrastive,gan2021skeleton,liu2026signpr,gan2025mixsigngraph,gan2026signllama,liu2026bridging,gan2026learning,gan2026sign}, including Isolated Sign Language Recognition (ISLR)~\cite{albanie2021bobsl}, Continuous Sign Language Recognition (CSLR)~\cite{koller2019weakly}, and Sign Language Translation (SLT)~\cite{gan2021skeleton,liu202}.
ISLR focuses on recognizing a single sign instance from a short, trimmed sign clip, typically formulated as a classification problem. CSLR aims to recognize unsegmented sign videos and output a sequence of gloss labels, addressing the challenge of temporal alignment between sign sequence and gloss sequence. In contrast, SLT~\cite{sincan2023context} goes beyond recognition by translating sign language (SL) sequences directly into spoken language sentences.
Despite strong semantic connections of these tasks, existing SLU methods are predominantly developed in a task-specific manner, where each task is addressed using an independent model architecture and training pipeline.
For ISLR, models usually adopt a visual backbone to capture spatiotemporal features, followed by a classification head for sign prediction.
For CSLR, a typical pipeline adopts a visual backbone for feature extraction~\cite{zhou2021spatial,gan2026signllama,gan23contrastive,gan2025mixsigngraph}, together with temporal modeling modules, and leverages CTC-based objectives to handle the latent alignment between  the sign videos and the gloss labels, yielding the target gloss sequence.
A similar task-specific design paradigm also dominates current SLT models. Most SLT approaches first employ a sign language backbone to extract visual representations~\cite{zhou2021spatial,gan2026learning,gan23contrastive,gan2025mixsigngraph}, followed by temporal encoding modules. The resulting sign features are then concatenated with textual embeddings and fed into a translation model and processed with the same transformer layer (\eg GPT-2~\cite{wong2024sign2gpt}, Gemma ~\cite{guo2025bridging}), to generate the corresponding spoken language sentence. 
  
\begin{figure}[t]
\centering
\includegraphics[width=0.99\linewidth]{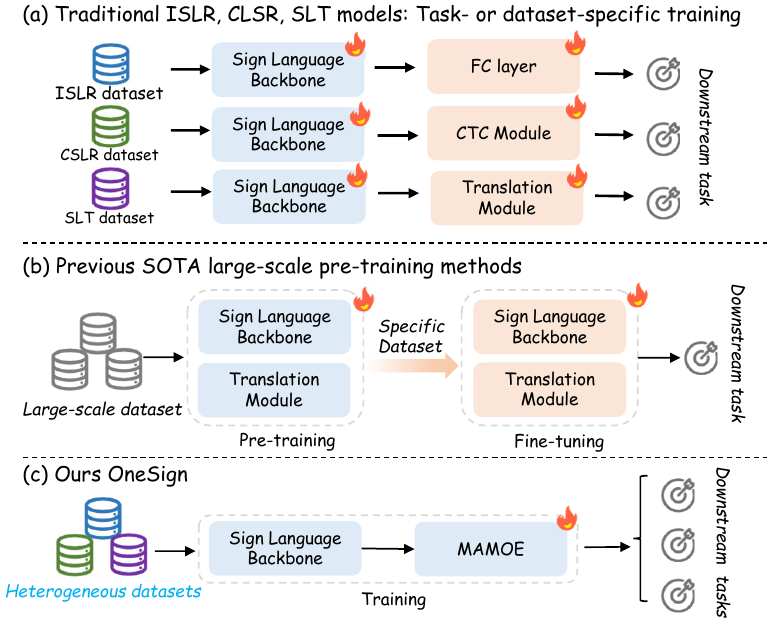} 
\caption{\label{fig:model1}Comparison between previous SLU pipelines and the proposed OneSign framework.}
\vspace{-5mm}
\end{figure}

Recently, several unified SL models (\eg MSLU~\cite{zhou2025scaling}, UniSign~\cite{li2025uni} and ShuBERT~\cite{gueuwouetal2025shubert}) have shown promising progress toward addressing multiple SLU tasks within a pretraining framework. These approaches typically rely on large-scale SL datasets for pretraining, followed by task- or dataset-specific finetuning to obtain dataset-dependent model checkpoints, achieving notable performance improvements. Despite their success, existing unified SLU models still exhibit two key limitations.  First, they depend on large annotated SL datasets for pretraining, which are expensive to construct. However, models still need to be finetuned separately for each downstream dataset, limiting training efficiency and model reusability. Second, many recent methods adopt decoder-only architectures, where large language models (LLMs) (\eg GPT-2 or Gemma) are used as translation modules. In these frameworks, sign tokens and text tokens are typically concatenated into a single sequence and processed by the same Transformer decoder layers, implicitly treating sign and text representations as belonging to the same modality. However, due to the inherent distributional differences between continuous sign representations and discrete textual tokens, this uniform modeling strategy overlooks the modality gap, which can hinder effective cross-modal reasoning and representation learning.

\input{table/CSLWB}

In this paper, we focus on core Sign Language Understanding (SLU) tasks, including ISLR, CSLR, and SLT, and propose OneSign, a unified framework that aims to solve heterogeneous SLU tasks within a single model and a single checkpoint. Unlike existing frameworks that follow a rigid pretraining–finetuning paradigm, where models are first pretrained on large-scale annotated datasets or carefully designed pretext tasks, and then separately finetuned on each downstream task. OneSign reformulates SLU training under a unified paradigm (See Figure~\ref{fig:model1}). This enables heterogeneous SLU datasets, including ISLR, CSLR, and SLT datasets, to be jointly trained within the same framework, without task-specific pretraining or retraining.

Furthermore, we observe that directly modeling sign tokens and text tokens within a standard LLM introduces significant challenges due to the inherent distributional gap between continuous sign representations and discrete textual tokens. To address this issue, we propose a Modality-Adaptive Mixture-of-Experts (MA-MoE) architecture, which explicitly accounts for modality heterogeneity while preserving shared semantic modeling. MA-MoE consists of an always-activated shared expert that captures modality-invariant sign–text semantics, together with modality-specific experts specialized for sign and text representations, respectively.

Through unified training and modality-adaptive modeling, OneSign effectively models the difference between heterogeneous sign and text modalities, enabling efficient multi-task learning and inference across diverse SLU tasks using a single model checkpoint (Figure.~\ref{fig:overall}).
The main contributions are as follows:
\begin{itemize}
    \item We construct \textbf{CSL-OpenWorld}, a large-scale Chinese SL (CSL) dataset consisting of over 100k video samples with manually curated annotations collected from diverse web sources across multiple domains (\eg news, law, and culture), substantially expanding existing CSL datasets. The dataset will be publicly released via HuggingFace. 
    
    \item We introduce a \textbf{Modality-Adaptive Mixture-of-Experts (MA-MoE)} architecture for LLM-based SLU modeling, which combines an always-activated shared expert with modality-specific experts to explicitly model the distributional difference between continuous sign representations and discrete text tokens.

    \item We propose \textbf{OneSign}, the first unified framework that reformulates ISLR, CSLR, and SLT under one learning paradigm, enabling unified training and inference across heterogeneous SLU tasks with a single model checkpoint, without task-specific retraining. Extensive experiments on multiple benchmarks demonstrate the effectiveness of OneSign across multiple SLU tasks.
\end{itemize}

%% file: table/CSLWB.tex
\begin{table}[!t] 
	\centering 
 \caption{Comparison with existing SLT datasets. \textbf{Vocab}: vocabulary size; \textbf{Hours}: total video duration in hours; \textbf{Topic}: main domain of videos (\eg daily life, TV news).} 
    \resizebox{0.97\linewidth}{!}{
    \begin{tabular}{lcccccc}
        \toprule
        Name & Language & Vocab. & Hours & Source & Topic \\
        \midrule
        KETI~\citep{ko2019neural} & KVK & 419 & 28 & Lab & Emergency Situations \\
         VRT-RAW~\citep{camgoz2021content4all} & VGT & - & 100 & TV &Broadcast \\
        SWISSTXT~\citep{camgoz2021content4all} & DGS & - & 88 & TV & News \\ 
        PHOENIX-2014T~\citep{cihan2018neural} & DGS & 3K & 11 & TV & Weather \\
        DGS Corpus~\citep{hanke2020extending} & DGS & - & 50 & Lab &Dialogue \\
        BOBSL~\citep{albanie2021bobsl} & BSL & 77K & 1,447 & TV & Broadcast  \\
        How2Sign~\citep{duarte2021how2sign} & ASL & 16K & 79 & Lab & Daily \\
        OpenASL~\citep{shi2022open} & ASL & 33K & 288 & Web & Multi-Topic \\
        YouTube-ASL~\citep{uthus2023youtube} & ASL & 60K & 984 & Web & Multi-Topic \\
        SP-10~\citep{yin2022mlslt} & Various & 17K & 14 & Web & Multilingual \\
        AfriSign~\citep{gueuwou2023afrisign} & Various & 20K & 152 & Web &Multilingual \\
        CSL-Daily~\citep{zhou2021improving} & CSL & 2K & 23 & Lab & Daily \\ 
        CSL-News~\cite{li2025uni} & CSL & 5K & 1,985 & TV & News \\
        \midrule
        CSL-OpenWorld (Ours) & CSL & 12K & 216 & Web & Multi-Topic \\
        \bottomrule 
    \end{tabular}}
\label{tab:chineseWB1}  
\end{table}  

%% file: Section/related.tex
 
\paragraph{ISLR.}  

SL Recognition (SLR) aims to recognize linguistic units from sign inputs and contains two tasks: ISLR and CSLR. Although both tasks focus on understanding sign semantics from videos, they differ primarily in supervision granularity and output structure, and are therefore typically treated as distinct problems. 
ISLR~\cite{hu2021hand, li2020transferring, zuo2023natural}~\nocite{liu2016sign, hu2021hand,hu2020global,uebersax2011real,zhao2023best,guo2016sign,liu2016sign,zhou2025scaling,zuo2023natural,li2020transferring,wong2025signrep} considers short, trimmed video clips containing a single sign and formulates recognition as a classification task.  The standard ISLR paradigm follows a task-specific pipeline, where a visual backbone extracts spatiotemporal features that are directly mapped to predefined sign categories via a classification layer. 
\paragraph{CSLR.}  
In contrast, CSLR~\cite{Zhang_2023_ICCV, gan2024signgraph, wei2023improving, hu2023self, jiao2023cosign} addresses the recognition of unsegmented sign sequences from continuous video streams, introducing the challenge of implicitly aligning visual inputs with gloss sequences. As a result, CSLR is commonly formulated as a sequence recognition problem with weak temporal supervision, where the input sign sequences and  gloss sequences share the same order but differ in length. To handle this alignment, the majority of CSLR methods adopt Connectionist Temporal Classification (CTC)-based frameworks. These approaches typically employ a vision backbone to extract sign features, followed by temporal modules to capture long-range dependencies, and a CTC-based decoder to align visual sequences with gloss sequences.

\paragraph{SLT.} Early SLT approaches are closely tied to advances in CSLR~\cite{wei2023improving,zuo2022c2slr,gan2024signgraph,niu2020stochastic,zhou2020spatial}~\nocite{wei2020semantic,cheng2020fully,zuo2022c2slr,pu2020boosting,hao2021self,min2021visual,zhang2019continuous,tang2021graph,yang2019sf,niu2020stochastic,Zhang_2023_ICCV,Hu_2023_CVPR,hu2022temporal,zhu2024multiscale,wei2023improving, hu2023self,parelli2022spatio,jiao2023cosign}. The prevailing paradigm in gloss-based SLT (GBSLT)~\cite{gan2023towards,yin2021simulslt,kan2022sign,jin2021contrastive,orbay2020neural,camgoz2020sign} relies on a pretrained CSLR backbone to provide intermediate gloss-aligned representations, which are subsequently fed into translation modules. In this setting, CSLR effectively serves as a pretext task that functions as a “sign tokenizer,” aligning visual features with gloss sequences and enabling downstream translation.
Further, increasing attention has been directed toward Gloss-Free SLT (GFSLT), which aims to directly translate sign videos into spoken language without intermediate gloss supervision. A fundamental challenge in GFSLT lies in how to tokenize sign sequence into discriminative sign features in the absence of gloss guidance. To address this issue, existing GFSLT methods typically rely on text annotations during pretraining, adopting strategies such as contrastive learning~\cite{zhou2023gloss,jiao2024visual,ye2024improving,liang2024llava}, CTC-based pseudo-gloss objectives~\cite{gan2025mixsigngraph}, text-aligned vector quantization~\cite{gong2024llms}, and multi-stage or curriculum-based pretraining schemes~\cite{guo2025bridging,gueuwouetal2025shubert,gueuwou2025signmusketeers}. These approaches aim to compensate for the lack of explicit gloss supervision by leveraging auxiliary alignment signals from textual modalities.

Existing SLR and SLT methods, though effective, are largely task-specific, with models and training objectives differing across ISLR, CSLR, and SLT. This fragmentation limits representation sharing and cross-task knowledge transfer, motivating the need for a unified, task-agnostic framework. Recent works, such as UniSign~\cite{li2025uni}, MSLU~\cite{zhou2025scaling}, and ShuBERT\cite{gueuwouetal2025shubert}, attempt multi-task learning by jointly pretraining their visual backbones or translation modules on large SL datasets and fine-tuning on each specific dataset.  However, these approaches still rely on task- or dataset-specific fine-tuning.

\paragraph{Mixture of Experts (MOE).}
Compared with standard LLMs that process all tokens using shared parameters, MoE~\cite{fedus2022switch,jiang2024mixtral,dai2024deepseekmoe} introduces  conditional computation by routing each input token to a subset of experts. This design enables a better trade-off between model capacity and computational efficiency, and has been shown to be particularly effective in handling heterogeneous data distributions and mitigating task interference~\cite{lepikhin2020gshard}. Existing MoE architectures can be broadly categorized into dense MoE and sparse MoE. In dense MoE~\cite{dou2024loramoe,pan2024dense}, all experts are activated during both training and inference, with different experts contributing via weighted combinations. In contrast, sparse MoE selects only Top-$k$ experts for each token, leading to significantly improved computational efficiency while maintaining strong model capacity.  In addition, prior works such as VLMo~\cite{vlmo} and BEiT-3~\cite{wang2023image} introduce mixture-of-modality-experts, which include a vision expert, a language expert, and a vision-language expert to handle different data pairs during vision-language pretraining. During fine-tuning, these models typically encode image and text separately.

Inspired by VLMo and BEiT-3, we propose MA-MoE, a sparse MoE architecture consisting of a shared expert and modality-specific experts for sign and text tokens. Compared with existing multimodal MoE models, MA-MoE differs in two key aspects.
(1) MA-MoE is specifically designed to address the modality heterogeneity in sign language understanding, which has been largely underexplored in prior SLU works.
(2) MA-MoE maintains a consistent architecture during both training and inference, whereas models such as VLMo and BEiT-3 require task-specific adaptations to the inference structure for different downstream tasks.


%% file: Section/method.tex
\begin{figure}
  \centering
 \includegraphics[width=0.44\linewidth]{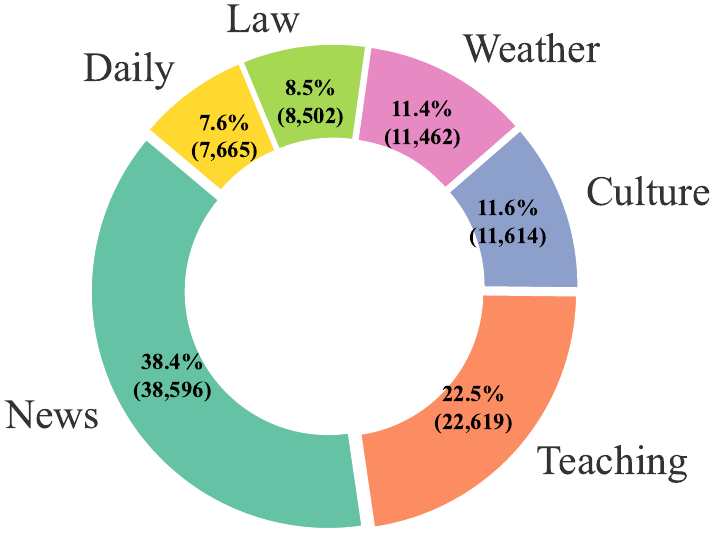}  
 \includegraphics[width=0.44\linewidth]{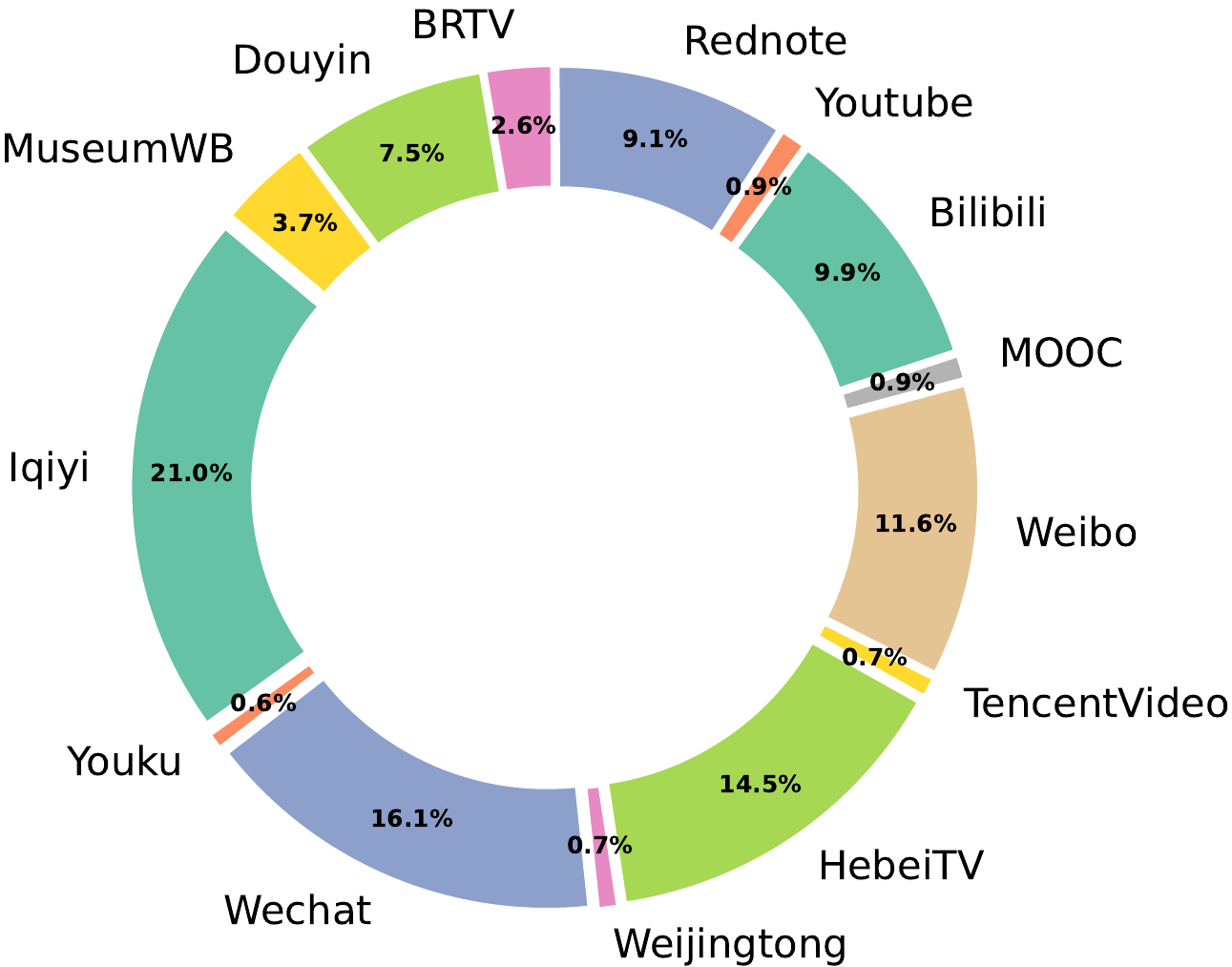} 
\caption{\label{fig:chineseWB2} Distribution of topics  and data sources.}
\vspace{-3mm}
\end{figure}

\subsection{Datasets}

The lack of large-scale, high-diversity corpora has long been a major bottleneck in SL research. While substantial progress has been made for American SL (ASL)~\cite{tanzer2025fleurs,uthus2023youtube}s and British SL (BSL)~\cite{albanie2021bobsl}~\nocite{tanzer2025fleurs,dinh2025sign,dinh2025sign,li2020word,desai2023asl,gueuwou2023afrisign,yin2022mlslt,uthus2023youtube,shi2022open,albanie2021bobsl,hanke2020extending,camgoz2021content4all,ko2019neural}, Chinese SL (CSL) remains under-resourced, limiting the generality and scalability of existing methods. Moreover, existing CSL benchmarks are largely “domain-locked,” consisting of either low-vocabulary laboratory recordings (\eg CSL-Daily) or highly formalized and linguistically constrained television broadcasts (\eg CSL-News). As a result, these CSL datasets fail to capture the morphological richness and environmental variability of natural signing.

\begin{figure}
  \centering
\includegraphics[width=0.99\linewidth]{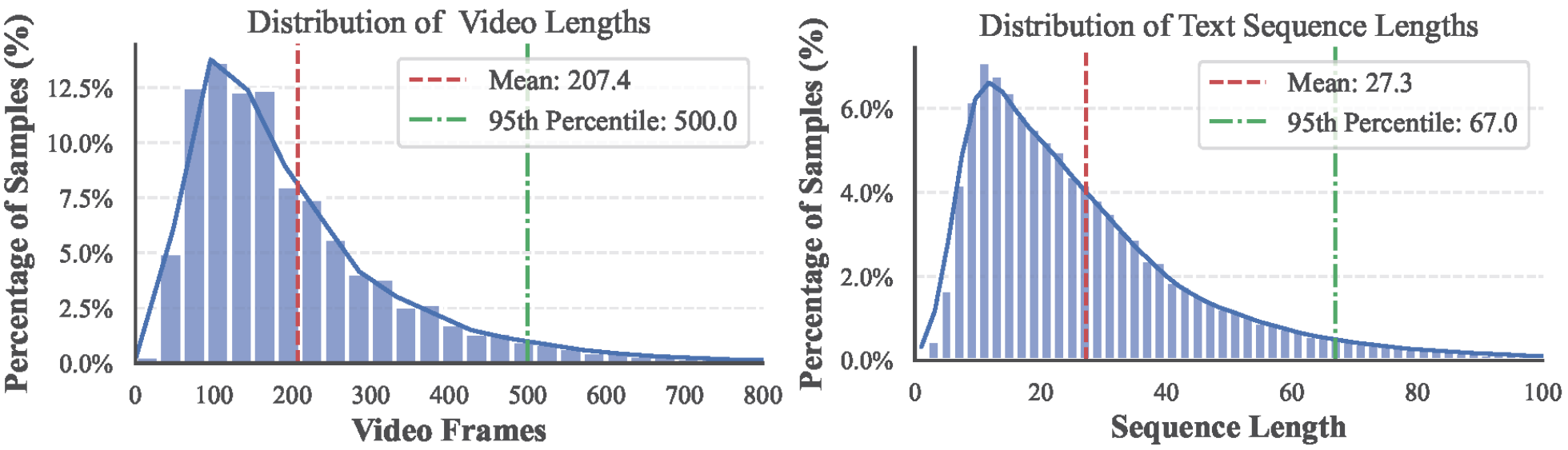}   
\caption{\label{fig:chineseWB3}Distribution of video durations and text lengths.}
\vspace{-3mm}
\end{figure}

To address these limitations, we introduce \textbf{CSL-OpenWorld}, a large-scale web-sourced dataset  for open-domain SLT. As summarized in Table~\ref{tab:chineseWB1}, CSL-OpenWorld expands the CSL vocabulary to over 12K tokens by leveraging diverse online platforms, representing a sixfold increase over CSL-Daily and more than that of CSL-News. Furthermore, as illustrated in Figures~\ref{fig:chineseWB2} and~\ref{fig:chineseWB3}, CSL-OpenWorld covers a wide range of topics, camera viewpoints, and non-studio environments, providing a rigorous testbed for evaluating model robustness and real-world generalization.

CSL-OpenWorld contains 100,458 video samples in total, with 86,387 used for training, making it substantially larger than CSL-Daily and Phoenix14t. The dataset features a high-density token vocabulary, with 12K unique tokens, enabling a challenging benchmark for evaluating the generalization ability of SLT models. The dataset is constructed through a rigorous pipeline involving web crawling, manual annotation, and quality filtering to ensure accurate video–text pairs. By incorporating diverse, open-domain web data, CSL-OpenWorld serves as a critical component in evaluating OneSign, enabling  assessment of model robustness across domains and linguistic variations, and providing a useful resource for advancing future SL research.

\begin{figure*}
    \centering
    \includegraphics[width=0.96\linewidth]{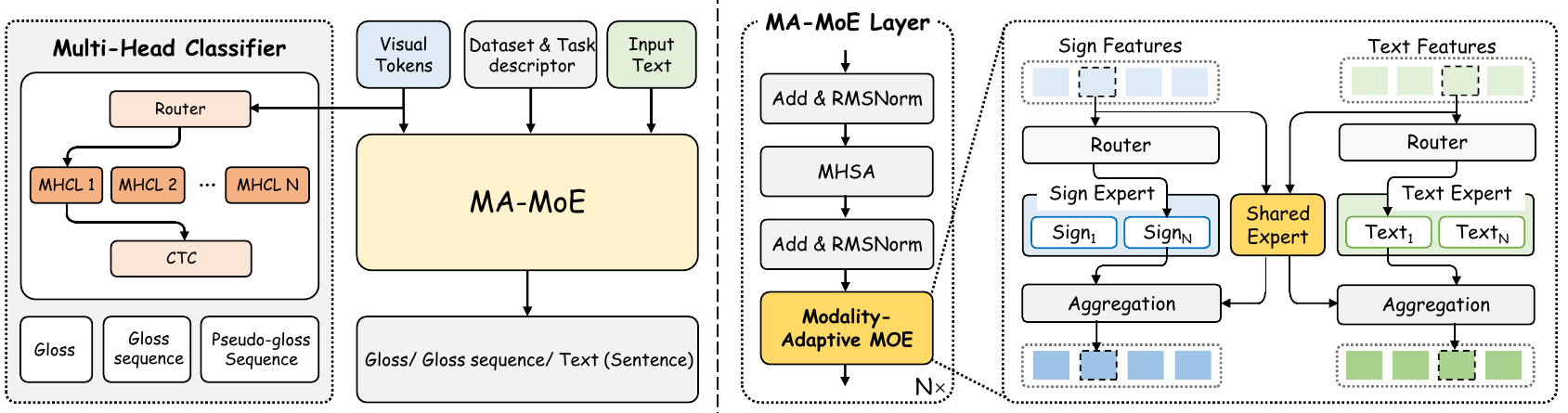}
\caption{Left: the OneSign Architecture. Right: The MA-MoE Layer with Modality Adaptive MoE structure.}
    \label{fig:model2} 
    \vspace{-3mm}
\end{figure*}

\subsection{Preliminaries.}
\noindent \textbf{ISLR task formulation.} ISLR focuses on a short sign clip containing a single sign, { formulated as } a classification problem. Given an input sign sequence $\mathbf{X}$, the objective is to find the optimal sign label $g^* \in \mathcal{G}$ that maximizes the expected probability under the model parameters $\theta$: 
\begin{equation}
    g^* = \arg\max_{g \in \mathcal{G}} \mathbb{E}_{g' \sim p_\theta(\cdot \mid \mathbf{X})} [\mathbb{I}(g = g')]
\end{equation}
where $\mathbb{I}(\cdot)$ denotes the indicator function, effectively seeking the Maximum A Posteriori (MAP) estimate.  

\noindent \textbf{CSLR task formulation.} CSLR aims to recognize continuous sign sequences by predicting a gloss sequence $\mathbf{G}=\{g_1,\dots,g_L\}$. Since the mapping involves a latent alignment between signs and glosses, the objective is formulated as the expectation over all valid align paths $\pi$ that map to $\mathbf{G}$ through a many-to-one function $\mathcal{B}$:
\begin{equation}
    \mathbf{G}^* = \arg\max_{\mathbf{G} \in \mathcal{G}^L} \mathbb{E}_{\pi \sim p_\theta(\cdot \mid \mathbf{X})} [\mathbb{I}(\mathcal{B}(\pi) = \mathbf{G})]
\end{equation} 

\noindent \textbf{SLT task formulation.}
SLT aims to translate sign sequences into a text sequence $\mathbf{Y}^* = \{y_1, \dots, y_U\}$. 
This task is commonly formulated as a conditional language generation problem, where the objective is to find $\mathbf{Y}^*$ that maximizes the expected log-likelihood (or  minimizes the Bayes risk) over the target space $\mathcal{Y}^U$, expressed as:
\begin{equation}
    \mathbf{Y}^* = \arg\max_{\mathbf{Y} \in \mathcal{Y}^U} \mathbb{E}_{\mathbf{Y}' \sim p_\theta(\cdot \mid \mathbf{X})} [\mathbb{I}(\mathbf{Y} = \mathbf{Y}')]
\end{equation}
In practice, this is often decomposed using the chain rule of probability as $p_\theta(\mathbf{Y} \mid \mathbf{X}) = \prod_{u=1}^{U} p_\theta(y_u \mid \mathbf{Y}_{<u}, \mathbf{X})$, where the model learns to predict the next token conditioned on $\mathbf{X}$ and the previously generated words $\mathbf{Y}_{<u}$.

\paragraph{Unified Perspective.} Despite  differences, all tasks can be unified under a single probabilistic framework.
In {OneSign}, each sample is represented as $\{\mathbf{X}, \mathbf{G}, \mathbf{Y}\}$. The model is trained by maximizing the joint conditional likelihood:
\begin{equation}
\mathcal{L}(\theta)
=
\mathbb{E}_{(\mathbf{X}, \mathbf{G}, \mathbf{Y}) \sim \mathcal{D}}
\left[
\log ( p_\theta(\mathbf{G},\mathbf{Y} \mid \mathbf{X}))
\right].
\end{equation}

This unified formulation allows {OneSign} to be trained under a single maximum-likelihood objective  from heterogeneous datasets. Moreover, it naturally subsumes multiple SLU tasks as special cases:
\begin{itemize}
\item \textbf{ISLR:} $\mathbf{G} = \mathbf{Y} = g$, where both denote a single sign label.
\item \textbf{CSLR:} $\mathbf{G} = \mathbf{Y} = \{g_1, \dots, g_L\}$, corresponding to a gloss sequence.
\item \textbf{GBSLT:} $\mathbf{G}$ denotes the gloss sequence, while $\mathbf{Y}$ represents the spoken language sentence.
\item \textbf{GFSLT:} $\mathbf{G}$ corresponds to a pseudo-gloss sequence, and $\mathbf{Y}$ denotes the spoken language sentence.
\end{itemize}

\subsection{Model Architecture}
As illustrated in Figure~\ref{fig:model2}, our model consists of three main components: 
(i) a visual encoder that converts an input sign sequence into a sequence of visual features, 
(ii) a text encoder that maps text tokens into continuous text representations, and 
(iii) a Modality-Adaptive Mixture-of-Experts (MA-MoE) decoder that performs unified sign-text modeling. To enable efficient training on large-scale datasets under limited GPU resources, we adopt pose-based representations instead of raw RGB videos for all tasks. 

Specifically, for a sign sequence  $\mathbf{X} = \{x_1, \dots, x_T\}$, each frame $x_t$ is represented using human pose information and decomposed into four semantic parts: 
$x_t = \{x_t^{b}, x_t^{f}, x_t^{lh}, x_t^{rh}\}$, corresponding to \emph{torso}, \emph{face}, \emph{left hand}, and \emph{right hand}, respectively. We first project each part into a shared high-dimensional feature space using independent linear transformations:
$\mathbf{h}_t^{p} = {W}_p x_t^{p}, \quad p \in \{b, f, lh, rh\},$
where ${W}_p$ denotes a learnable linear projection. This yields four part-level feature embeddings per frame.

\paragraph{Dynamic Graph Fusion for Body Part.}To effectively model the spatial correlations and structural dependencies between these body parts, inspired by previous pose modeling~\cite{shi2019two, ye2020dynamic, chen2021channel}, we propose a simple yet effective pose fusing method: the Dynamic Part Graph Fusion (DPGF). Unlike static graph models, DPGF treats each body part as a node in a dynamic graph and learns the underlying topology on-the-fly.
Specifically, sign features are concatenated to form a node matrix $\mathbf{H} \in \mathbb{R}^{4 \times D}$, where each  node represents a body part.  We employ a self-attention mechanism to compute the dynamic adjacency matrix $\mathbf{A} \in \mathbb{R}^{4 \times 4}$, which defines the edge strengths between parts for each frame:
\begin{equation} 
    \mathbf{A} = \text{Softmax} \left( \frac{Q(\mathbf{H}) K(\mathbf{H})^\top}{\sqrt{D}} \right) 
\end{equation}
where $Q$ and $K$ are linear transformations for Query and Key. This allows the model to adaptively focus on different part correlations (\eg hand-face coordination) depending on the sign being performed.  The node features are updated through a weighted aggregation of their neighbors followed by a GCN-style update with a residual connection:
\begin{equation}
    \mathbf{H}_{fused} = W* (\text{MLP}(\mathbf{A} V(\mathbf{H})) + \mathbf{H})
\end{equation}
where $V$ is the linear projection. The updated nodes are then flattened and projected  with $W$ to the final feature space $\mathbf{H}_{fused} \in \mathbb{R}^{B \times T \times C}$, providing a comprehensive representation of the sign gesture for each frame.

\paragraph{Multi-Head Classifier Layers.} 
The visual backbone is trained with a CTC objective using $\mathbf{G}$ as supervision. When mixing multiple heterogeneous SL datasets, directly using a single shared classifier would require merging their gloss vocabularies, resulting in an excessively large output space and severe label sparsity. Inspired by previous multi-task learning~\cite{liu2015representation,ruder2017overview,long2017learning}, we therefore adopt {Multi-Head Classifier Layers (MHCL)}, in which each dataset is assigned a dedicated classification head with its own gloss vocabulary, while all heads share the same visual backbone. This strategy allows the backbone to learn unified visual representations across datasets, while each classifier head specializes in its dataset-specific label space.
By decoupling feature learning from dataset-specific vocabularies, this design improves optimization stability and enables scalable multi-dataset pretraining. \textit{MHCL is only used during training to stabilize optimization. It is removed at inference, where all outputs are generated by the unified translation module without dataset-specific heads.}

\subsection{Modality-Adaptive LLM MoE}
After obtaining sign features, many  approaches concatenate sign and text tokens into a single sequence and process them using standard LLM layers. Such a design implicitly treats sign and text tokens as the same modality, overlooking their inherent modality-specific characteristics and relying entirely on shared parameters across all decoder layers. 

However, this design relies on an implicit assumption: \emph{sign tokens and textual tokens are well aligned in the latent space}, such that a unified set of transformation layers can model both modalities equally well without explicit distinction. In practice, sign representations derived from pose sequences exhibit markedly different statistical properties, temporal dynamics, and levels of semantic abstraction compared to discrete text tokens. As a result, forcing both modalities to share identical transformation paths introduces a non-negligible \emph{modality gap}, which can hinder effective cross-modal modeling and degrade overall performance.

To address this issue, we propose a  \textbf{Modality-Adaptive Mixture-of-Experts (MA-MoE)} architecture that explicitly models modality-specific distributions while preserving shared semantic representations. As illustrated in Fig.~\ref{fig:model2}, MA-MoE augments a standard Transformer block by introducing modality-adaptive expert routing within the feed-forward network (FFN). Specifically, the expert pool is decomposed into three disjoint groups: a shared expert, $N$ sign-specific experts, and $M$ text-specific experts. The shared expert is always activated to capture modality-invariant semantics, while modality-specific experts focus on modeling characteristics unique to sign or text tokens.

Let the input to the MA-MoE layer be a sequence of token representations: $\mathbf{H} = \{\mathbf{h}_1, \dots, \mathbf{h}_N\}, \quad \mathbf{h}_i \in \mathbb{R}^D$, where each token is associated with a modality label
$m_i \in \{\text{sign}, \text{text}\}$.
The MA-MoE module consists of a shared expert $\mathcal{E}_{\text{shared}} = \{E_s\}$,
a set of sign-specific experts $\mathcal{E}_{\text{sign}}$,
and a set of text-specific experts $\mathcal{E}_{\text{text}}$. For each token $\mathbf{h}_i$, the shared expert output is first computed as
\begin{equation} 
\mathbf{h}_i^{s} = E_s(\mathbf{h}_i). 
\end{equation}
Modality-specific routing is then performed within each expert group. A lightweight router produces routing logits:
\begin{equation}
\mathbf{r}_i =
\begin{cases}
\mathrm{Router}_{\text{sign}}(\mathbf{h}_i), & \text{if } m_i = \text{sign}, \\
\mathrm{Router}_{\text{text}}(\mathbf{h}_i), & \text{if } m_i = \text{text},
\end{cases}
\end{equation}
which are normalized over the modality-specific expert set:
\begin{equation}
\alpha_{i,e} =
\frac{\exp(r_{i,e})}
{\sum\limits_{e' \in \mathcal{E}_{m_i}} \exp(r_{i,e'})},
\quad e \in \mathcal{E}_{m_i}. 
\end{equation}
We denote the set of selected top-$k$ expert indices as $\tau$. The final output representation is obtained by aggregating the shared and modality-specific expert outputs:
\begin{equation}
\tilde{\mathbf{h}}_i =
\mathbf{h}_i^{s}
+ \sum_{e \in \tau} \alpha_{i,e} \, \mathcal{E}_{m_i, e}(\mathbf{h}_i) 
\end{equation}
By construction, MA-MoE disentangles modality-invariant alignment from modality-specific modeling, enabling effective joint reasoning over heterogeneous sign and text representations within a unified LLM architecture.

\paragraph{Load Balancing Loss.}
To prevent expert collapse and encourage uniform utilization of modality-specific experts, we introduce a load balancing loss for the MA-MoE.
Since the shared expert is always activated, the balancing constraint is applied only to modality-specific experts. For a given modality $m \in \{\text{sign}, \text{text}\}$, let $\mathcal{E}_m$ denote its corresponding set of modality-specific experts, and let $N_m$ be the number of tokens of modality $m$ in a mini-batch. Given the routing probabilities $\alpha_{i,e}$ for token $\mathbf{h}_i$ and expert $e \in \mathcal{E}_m$, we define the importance of expert $e$ as:
\begin{equation}
\mathrm{Imp}(e) = \frac{1}{N_m} \sum_{i: m_i = m} \alpha_{i,e}
\end{equation}

The load balancing loss for modality $m$ is  formulated as:
\begin{equation}
\small
\mathcal{L}_{\mathrm{bal}}^{(m)}=|\mathcal{E}_m|\sum_{e \in \mathcal{E}_m}
\left(\mathrm{Imp}(e) - \frac{1}{|\mathcal{E}_m|}\right)^2
\end{equation}

The final load balancing loss is computed as the weighted sum over modalities:
\begin{equation}
\mathcal{L}_{\mathrm{bal}}
=
\lambda_{\text{sign}} \, \mathcal{L}_{\mathrm{bal}}^{(\text{sign})}
+
\lambda_{\text{text}} \, \mathcal{L}_{\mathrm{bal}}^{(\text{text})}
\end{equation}
where $\lambda_{\text{sign}}$ and $\lambda_{\text{text}}$ are hyperparameters controlling the strength of the balancing constraint for each modality. This loss encourages the router to distribute tokens evenly across modality-specific experts while still allowing experts to specialize in capturing modality-dependent patterns. As a result, the load balancing loss improves training stability and ensures effective capacity utilization within each modality.

\subsection{Training}
The OneSign model is optimized with two objectives that combine recognition and translation losses:
\begin{equation}
\resizebox{0.85\linewidth}{!}{%
$ 
\begin{aligned}
\mathcal{J}(\Theta)
&=\;\alpha \,
\mathbb{E}_{(X, G) \sim \mathcal{D}}
\left[ \ell_{\mathrm{CTC}}\!\left( R(X; \Theta_R), G \right) \right] \\
&+ \beta \,
\mathbb{E}_{(\tilde{\mathbf{h}}_i, Y) \sim \mathcal{D}}
\Big[
\ell_{\mathrm{CE}}\!\left( Y, T(\tilde{\mathbf{h}}_i) \right)
+ \lambda_{s} \mathcal{L}_{\mathrm{bal}}^{(s)}
+ \lambda_{t} \mathcal{L}_{\mathrm{bal}}^{(t)}
\Big].
\end{aligned}
$ 
}
\label{equ:total_objective}
\end{equation}
Here, $R$ and $T$ denote the recognition and translation modules, respectively.
Following~\cite{gan2025mixsigngraph}, the CTC loss supervises the visual backbone with MHCL, while the translation module $T$ is optimized with a translation loss with MOE balancing losses. During training, we first warm up the visual backbone for 20 epochs by setting $\alpha=1$,$\beta=0$, and activate the translation objective by further setting $\beta=1$ in the subsequent 30 epochs.

%% file: Section/performance.tex
\input{table/expert}

\input{table/expert1_2}

\section{Ablation Study}\label{sec:ablation}
Following previous work~\cite{cihan2018neural, chen2022two, gan2024signgraph}, we perform ablation studies on the Phoenix14t dataset to verify the effectiveness of the proposed OneSign model. \textit{Please refer to Appendix~\ref{sec:Imp} for detailed implementation settings.}
\paragraph{A1: Effect of the Multi-Head Classification Layer.}
Using a single CTC-based classification head for heterogeneous SLU datasets leads to a large vocabulary and unstable training. Table~\ref{tab:MHCL} compares three settings: (1) \textbf{OneSign$^\heartsuit$} , trained on a single dataset; (2) \textbf{OneSign$^\spadesuit$}  W/O MHCL, trained on multiple datasets with a shared head; and (3) \textbf{OneSign$^\spadesuit$} with an independent head per dataset. As can be seen, directly sharing a head across datasets dilutes the original distributions and yields limited gains, whereas MHCL improves both performance and training stability, highlighting its effectiveness for multi-dataset learning.

\input{table/expert2}

\input{table/expert2_2}

\input{table/dpgf}

\input{table/time}

\paragraph{A2: Effect of Heterogeneous Training Datasets.}
Table~\ref{tab:TrainingDataset} shows the impact of jointly training  on heterogeneous datasets: \emph{ISLR} (WLASL, ASLCitizen, MultiVSL), \emph{CSLR} (CSL-Daily, Phoenix14t), and \emph{SLT} (CSL-Daily, Phoenix14t, How2Sign, CSL-OpenWorld), evaluated on Phoenix14t. Adding datasets without task-specific finetuning yields only modest gains. The results show that naively adding heterogeneous datasets brings only modest improvements (\eg +CSLR to +ISLR+CSLR), as the lack of any downstream finetuning can dilute the original dataset distributions. Still, OneSign effectively leverages complementary supervision, demonstrating the benefit of cross-task knowledge transfer in unified multi-dataset training.

\paragraph{A3: Effect of the Proposed MA-MoE.}
Table~\ref{tab:Expert} presents an ablation study on different expert configurations in the proposed MA-MoE. Using only the shared expert (\ie Llama3.2 1B model) results in suboptimal performance, indicating that modality-agnostic modeling alone is insufficient. Adding either a sign or text specific expert on top of the shared expert consistently improves performance, highlighting the value of modality-aware MOE design. The full configuration, which includes both sign  and text specific experts, achieves the best results across all metrics, confirming that MA-MoE effectively balances shared semantic modeling with modality-specific representation learning.

\paragraph{A4: Effect of Sparse MA-MoE Configurations.} 
Table~\ref{tab:Expert2} examines expert sparsity by varying the total number of experts ($N_s=N_t$) and activated experts (Top-$k$). Activating a single expert with more total experts gives minor gains, while activating multiple experts allows aggregation of complementary information, but requires more computational resources. To balance performance and efficiency, we choose $N_s=N_t=2$ with Top-$k=1$ as our final setting.

\paragraph{A5: Effect of Dynamic Part Graph Fusion.}
Table~\ref{tab:DGFP} reports an ablation study on the proposed Dynamic Part Graph Fusion (DPGF) module.
When DPGF is removed (\ie features from different parts are fused using a simple linear operation), the model performs worse on all metrics, indicating that naive part fusion is insufficient to capture fine-grained part-level interactions.
Applying a fixed fusion strategy (W/ Fix), where fusion graphs are predefined for each part, brings modest and consistent improvements, suggesting that part-aware aggregation is beneficial but limited by its static design.
In contrast, our model with DPGF module achieves the best performance , reducing WER from 21.1 to 20.1 and improving BLEU4 from 25.63 to 26.45.
These gains indicate that dynamically adjusting fusion weights according to input content enables more effective modeling of inter-part dependencies, which in turn leads to more accurate recognition and translation.

\input{table/islr}

\paragraph{A6: Training and Inference Speed}

We evaluate the training and testing time on the Phoenix14t using a single NVIDIA RTX 4090 GPU. Specifically, training time is measured on the Phoenix14t training set, while inference speed is evaluated on test samples with an average length of 250 frames, averaged over 100 runs. As observed, the model requires about 10–13 minutes per epoch during training. During inference, it achieves a throughput of 10–12 samples per second on a single GPU. These results show that the proposed MA-MOE achieves good efficiency in both training and inference, making it practical for large-scale SL tasks.

\section{Comparisons}

We compare \textbf{OneSign} with existing task-specific  models on ISLR, CSLR, and SLT benchmarks. 
To systematically evaluate the effect of unified training and inference, we report results under two training and evaluation settings: (1) \textbf{OneSign$^\heartsuit$} is trained and evaluated on a single dataset(\ie each corresponding dataset).
(2) \textbf{OneSign$^\spadesuit$} is trained jointly on the training splits of all available datasets and directly evaluated on each benchmark without any further fine-tuning, measuring the generalization capability of a single unified checkpoint.\textit{Note that a direct comparison with RGB-based models is not entirely fair, as RGB inputs contain richer visual information.}

 \paragraph{Evaluation on ISLR Tasks.} 
Table~\ref{tab:ISLRCom} compares {OneSign} with existing ISLR methods. Prior approaches are mostly RGB- or pose-based and often require dataset-specific pretraining or fine-tuning.  Under the dataset-specific setting, {OneSign$^\heartsuit$} already show promising performance across all benchmarks (\eg 65.41 Top@1 on ASL Citizen vs. 59.52 for ST-GCN). On WLASL2000 and MultiVSL1000, it similarly achieves clear gains. {OneSign$^\spadesuit$} further improves performance, establishing a new baseline under a unified protocol. These results show that OneSign effectively leverages heterogeneous supervision and achieves strong  multi-dataset performance with a single checkpoint.

\input{table/islr_2}
\input{table/slt}

\paragraph{Evaluation on CSLR Tasks}  
Table~\ref{tab:CSLRCom} shows results on CSL-Daily and Phoenix14t. Under dataset-specific settings, {OneSign$^\heartsuit$} achieves strong performance using only pose inputs (29.1 WER on CSL-Daily, 22.1 on Phoenix14t), matching or surpassing several RGB-based methods.  {OneSign$^\spadesuit$}, trained jointly on heterogeneous datasets without fine-tuning, further reduces WER to 25.7 and 20.1, respectively, demonstrating good performance. Compared with prior pose-based methods that require fine-tuning, OneSign$^\spadesuit$ achieves comparable or better results under a unified evaluation setting.

\paragraph{Evaluation on SLT.}

Tables~\ref{tab:GBSLT} and~\ref{tab:GFSLT} report results on two gloss-based SLT benchmarks: Phoenix14t, CSL-Daily, and two gloss-free SLT benchmarks: How2Sign, and CSL-OpenWorld. Unlike most existing SLT methods that rely on  dataset-specific fine-tuning, OneSign operates purely on pose and supports unified training without fine-tuning. On Phoenix14t and CSL-Daily (Table~\ref{tab:GBSLT}), \textbf{OneSign$^\heartsuit$} already achieves competitive performance using only pose features.
Joint training further brings consistent gains: \textbf{OneSign$^\spadesuit$} improves BLEU4 from 26.22 to 26.45 on Phoenix14t and from 24.42 to 26.15 on CSL-Daily. On the more challenging How2Sign and CSL-OpenWorld benchmarks (Table~\ref{tab:GFSLT}), OneSign shows strong generalization.
Without fine-tuning, \textbf{OneSign$^\spadesuit$} achieves the best or highly competitive results among pose-based methods,  outperforming prior approaches such as Uni-Sign and VAP on many metrics. Overall, OneSign achieves strong cross-dataset performance with a single unified checkpoint, without relying on RGB inputs or dataset-specific fine-tuning.

%% file: table/expert.tex
\begin{table}[!t] 
	\centering 
 \captionof{table}{\label{tab:MHCL}Effect of Multi-Head Classification Layer.} 
	 \resizebox{0.95\linewidth}{!}{
\begin{tabular}{l|c@{\hspace{0.5em}}c|cc} 
\toprule
\multirow{2}*{Model}&  \multicolumn{2}{c}{DEV}&\multicolumn{2}{c}{TEST} \\
& WER& del/ins& WER& del/ins \\
\shline  
OneSign$^\heartsuit$ &{21.1}&7.2/3.0& {22.1}& 8.1/2.3 \\
OneSign$^\spadesuit$ W/O MHCL  &{23.5}&11.8/2.1 &{24.9} &11.7/2.5\\
OneSign$^\spadesuit$ &\baseline{20.0}& 5.2/3.1& \baseline{20.1} &5.7/2.9\\	
            \bottomrule
          \end{tabular}} 
\end{table}

%% file: table/expert1_2.tex
\setcounter{table}{2}
\begin{table}[!t] 
	\centering 
    \captionof{table}{\label{tab:TrainingDataset}Effect of heterogeneous training  datasets.} 
        \resizebox{0.9\linewidth}{!}{
    \begin{tabular}{l|cccc} 
            \toprule
       \multirow{2}*{Dataset}&\multicolumn{4}{c}{TEST} \\
   &WER & Rouge& BLEU1&BLEU4 \\
   \shline  
Phoenxi14T & 22.1&{51.18}&{52.68}&{26.22} \\
+CSLR& 21.5 &{51.65}&{52.45}&{26.25} \\
+ISLR+CSLR & 21.0 &{52.18}&{52.65}&{26.24}\\
+ISLR+CSLR+SLT & \baseline{20.1}&  \baseline{52.01}&\baseline{53.36}&\baseline{26.45}\\  	
        \bottomrule 
      \end{tabular}} 
\end{table}

%% file: table/expert2.tex
\setcounter{table}{3}
\begin{table}[!t] 
	\centering 
   \captionof{table}{\label{tab:Expert}Effect of MA-MOE.}  
       \tablestyle{1.2pt}{1.1} 
\resizebox{1.0\columnwidth}{!}{ 
		\begin{tabular}{ccc|cccc|cccc}
		\shline
  \multicolumn{3}{c|}{Experts}&  \multicolumn{4}{c|}{DEV} &  \multicolumn{4}{c}{TEST} \\
 Shared &Sign&Text &WER&Rouge& BLEU1& BLEU4&WER&Rouge& BLEU1& BLEU4 \\
\shline  
\checkmark &  & &22.3&49.12&49.89&23.45&{22.5}&  {48.99}&{49.72}&{23.01}\\
\checkmark&\checkmark  && 21.0&50.12&49.65&23.65&{21.4}&  {50.01}&{49.41}&{23.47}\\
 \checkmark& &\checkmark& 20.8&50.64&50.12&24.36&{21.0}&  {50.03}&{49.95}&{24.13}\\
 \checkmark& \checkmark &\checkmark &\baseline{20.0}& \baseline{52.34}&\baseline{53.41}& \baseline{26.98}&\baseline{20.1}&  \baseline{52.01}&\baseline{53.36}&\baseline{26.45}\\	 	
	\shline 
	\end{tabular} }
\end{table}

%% file: table/expert2_2.tex
\setcounter{table}{4}
\begin{table}[!t] 
	\centering  
       \captionof{table}{\label{tab:Expert2} {Impact of activated expert count.}}  
    \tablestyle{1.7pt}{1.0} 
        \resizebox{1.0\linewidth}{!}{
\begin{tabular}{cc|rr|ccc}
		\shline
$N_s=N_t$ & Top-$k$ & Params (M) & Activated (M)& Dev-Rouge & {Dev-BLEU1 }& Dev-BLEU4\\
	\shline 
                0 & 0
                & 1235.8 & 1235.8
                & 49.45 & 49.36 & 22.67 \\

                2 & 1
                & 4457.0 & 2041.1
                & 52.34 & 53.41 & 26.98 \\

                4 & 1
                & 7678.2 & 2041.1
                & 52.74 & 53.65 & 26.78 \\

                4 & 2
                & 7678.2 & 3651.7
                & 52.99 & 53.94 & 26.97 \\
	\shline 
\end{tabular}
}
\end{table}

%% file: table/dpgf.tex
\begin{table}[!t] 
	\centering 
  \caption{Effects of Dynamic Part Graph Fusion.}  
    \tablestyle{1.7pt}{1.0} 
\resizebox{0.95\linewidth}{!}{ 
\begin{tabular}{cc|cccc|cccc}
\toprule
&&\multicolumn{4}{c|}{Dev}&\multicolumn{4}{c}{Test} \\
&&WER&ROUGE&BLEU1&BLEU4&WER&ROUGE&BLEU1&BLEU4\\
\shline
W/O DPGF&& {21.6}&  {51.01}&{52.87}&{25.83}& {21.1}&  {51.07}&{52.17}&{25.63}\\
W/ Fix && {20.6}&  {51.87}&{52.36}&{26.36}&{20.7}&  {51.46}&{52.48}&{25.87}\\
W/ DPGF &&\baseline{20.0}&\baseline{52.35}&\baseline{53.41}&\baseline{26.98}& \baseline{20.1}&  \baseline{52.01}&\baseline{53.36}&\baseline{26.45}\\
\bottomrule
\end{tabular}}
\label{tab:DGFP}  
\end{table}

%% file: table/time.tex
\begin{table}[t]
\centering
\caption{Training speed and Inference speed.}
\resizebox{0.9\columnwidth}{!}{
\begin{tabular}{lcc} 
\toprule
Model & Training Time &Inference Time (250 frames) \\
\midrule
& $\sim$10.4 min/epoch
& $\sim$12.5 samples/s \\
+MA-MOE
& $\sim$11.2 min/epoch
& $\sim$11.8 samples/s \\
\bottomrule
\end{tabular}}  
\end{table}

%% file: table/islr.tex
\begin{table}[!t] 
\centering 
\captionof{table}{
\label{tab:ISLRCom}
ISLR results on various benchmarks.
$^*$ denotes results reproduced by us using publicly available code,
and $\checkmark^{*}$ indicates pretraining on large-scale SL datasets. }
\resizebox{1.0\linewidth}{!}{
\begin{tabular}{l|c@{\hspace{0.5em}}cc|cc|cc|cc} 
\toprule
{\multirow{2}{*}{{Method}}}
& \multicolumn{3}{c|}{{Extra}}
& \multicolumn{2}{c|}{{ASL Citizen}}
& \multicolumn{2}{c|}{{WLASL2000}}
& \multicolumn{2}{c}{{MultiVSL1000}}\\  
& Pose & RGB & FT
& Top@1 & Top@5
& P-I & P-C
& Top@1 & Top@5 \\
\midrule 

\gc{I3D}
& \gc{} & \gc{\checkmark} & \gc{\checkmark}
& \gc{63.10} & \gc{86.09}
& \gc{37.91} & \gc{35.90}
& \gc{40.60} & \gc{73.03} \\

\gc{HMA~\citep{hu2021hand}}
& \gc{} & \gc{\checkmark} & \gc{\checkmark}
& \gc{-} & \gc{-}
& \gc{37.91} & \gc{35.90}
& \gc{-} & \gc{-} \\

\gc{NLA-SLR~\citep{zuo2023natural}}
& \gc{} & \gc{\checkmark} & \gc{\checkmark}
& \gc{-} & \gc{-}
& \gc{61.05} & \gc{58.05}
& \gc{-} & \gc{-} \\ 

\gc{Uni-Sign~\cite{li2025uni}}
& \gc{\checkmark} & \gc{\checkmark} & \gc{\checkmark$^{*}$}
& \gc{-} & \gc{-}
& \gc{63.52} & \gc{61.32}
& \gc{-} & \gc{-} \\ 

\gc{SHuBERT~\cite{gueuwouetal2025shubert}}
& \gc{} & \gc{\checkmark} & \gc{\checkmark$^{*}$}
& \gc{65.00} & \gc{87.00}
& \gc{60.90} & \gc{58.01}
& \gc{-} & \gc{-} \\ 

\midrule
\rowcolor[gray]{.85}
\multicolumn{10}{c}{Pose-based}\\
\midrule 

ST-GCN~\citep{yu2017spatio}
& \checkmark & & \checkmark
& 59.52 & 82.68
& 34.40 & 32.53
& {40.61*} & {73.04*} \\  

BEST~\citep{zhao2023best}
& \checkmark & & \checkmark
& - & -
& 46.25 & 43.52
& - & - \\

SignBERT+~\citep{hu2023signbert+}
& \checkmark & \checkmark & \checkmark
& - & -
& 48.85 & 46.37
& - & - \\

MSLU~\citep{zhou2025scaling}
& \checkmark & & \checkmark$^{*}$
& - & -
& 56.29 & 53.29
& - & - \\  

Uni-Sign~\cite{li2025uni}
& \checkmark & & \checkmark$^{*}$
& {60.92*} & {83.41*}
& \underline{63.13} & 60.90
& {70.92*} & {89.90*} \\ 

\hline

OneSign$^\heartsuit$
& \checkmark & & \checkmark
& \baseline{65.41} & \baseline{84.19}
& \baseline{60.45} & \baseline{61.02}
& \baseline{80.71} & \baseline{95.43} \\ 

OneSign$^\spadesuit$
& \checkmark & &
& \baseline{68.41} & \baseline{89.24}
& \baseline{64.15} & \baseline{65.34}
& \baseline{82.34} & \baseline{96.43} \\ 

\bottomrule
\end{tabular}} 
\end{table}

%% file: table/islr_2.tex
\begin{table}[!t] 
\centering  
\captionof{table}{
\label{tab:CSLRCom}
CSLR results on CSL-Daily and Phoenix14T.
``FT'' denotes fine-tuning on each dataset.
}
\resizebox{1.0\linewidth}{!}{
\begin{tabular}{l|c@{\hspace{0.5em}}cc|cc|cc} 
\toprule
{\multirow{2}{*}{{Method}}}
& \multicolumn{3}{c|}{{Extra}}
& \multicolumn{2}{c|}{{CSL-Daily}}
& \multicolumn{2}{c}{{Phoenix14T}}\\   
& Pose & RGB & FT & Dev & Test & Dev & Test \\
\midrule

\gc{SignBT~\citep{zhou2021improving}}
& \gc{} & \gc{\checkmark} & \gc{\checkmark}
& \gc{33.2} & \gc{33.2}
& \gc{-} & \gc{-} \\

\gc{CorrNet~\citep{Hu_2023_CVPR}}
& \gc{} & \gc{\checkmark} & \gc{\checkmark}
& \gc{30.6} & \gc{30.1}
& \gc{18.9} & \gc{20.5} \\

\gc{SEN~\citep{hu2023self}}
& \gc{} & \gc{\checkmark} & \gc{\checkmark}
& \gc{31.1} & \gc{30.7}
& \gc{19.3} & \gc{20.7} \\

\gc{C2ST~\citep{Zhang_2023_ICCV}}
& \gc{} & \gc{\checkmark} & \gc{\checkmark}
& \gc{25.9} & \gc{25.8}
& \gc{17.5} & \gc{17.7} \\

\gc{SignGraph~\cite{gan2024signgraph}}
& \gc{} & \gc{\checkmark} & \gc{\checkmark}
& \gc{27.3} & \gc{26.4}
& \gc{17.8} & \gc{19.1} \\

\gc{Uni-Sign~\cite{li2025uni}}
& \gc{\checkmark} & \gc{\checkmark} & \gc{\checkmark$^{*}$}
& \gc{26.7} & \gc{26.0}
& \gc{-} & \gc{-} \\

\gc{TwoStream~\cite{chen2022two}}
& \gc{\checkmark} & \gc{\checkmark} & \gc{\checkmark}
& \gc{25.4} & \gc{25.3}
& \gc{17.7} & \gc{19.3} \\

\midrule
\rowcolor[gray]{.85}
\multicolumn{8}{c}{Pose-based}\\
\midrule  

CoSign1s~\citep{jiao2023cosign}
& \checkmark & & \checkmark
& 29.5 & 29.1
& 20.4 & \underline{20.6} \\

MSLU~\citep{zhou2025scaling}
& \checkmark & & \checkmark
& 28.6 & 27.9
& \underline{20.1} & 21.3 \\ 

Uni-Sign~\cite{li2025uni}
& \checkmark & & \checkmark$^{*}$
& \underline{28.2} & \underline{27.4}
& {21.4*} & {22.0*} \\ 

\midrule

OneSign$^\heartsuit$
& \checkmark & & \checkmark
& \baseline{29.0} & \baseline{29.1}
& \baseline{21.1} & \baseline{22.1} \\  

OneSign$^\spadesuit$
& \checkmark & &
& \baseline{25.2} & \baseline{25.7}
& \baseline{20.0} & \baseline{20.1} \\ 

\bottomrule 
\end{tabular}
} 
\end{table} 

%% file: table/SLT.tex
\begin{table*}[t] 
\centering 
\caption{
Comparison of SLT performance on Phoenix14T and CSL-Daily.
$^{*}$ denotes results reproduced by us using publicly available code.
We underline pose-based results that outperform our baseline model for clarity.
}
\label{tab:GBSLT}
\resizebox{0.97\textwidth}{!}{
\begin{tabular}{l|ccc|lll|lll|lll|lll} 
\toprule
\multirow{3}{*}{SLT}
& \multicolumn{3}{c|}{Extra}
& \multicolumn{6}{c|}{Phoenix14T}
& \multicolumn{6}{c}{CSL-Daily} \\
& & & 
& \multicolumn{3}{c|}{DEV}
& \multicolumn{3}{c|}{TEST}
& \multicolumn{3}{c|}{DEV}
& \multicolumn{3}{c}{TEST} \\ 
& Pose & RGB & FT
& ROUGE & BLEU1 & BLEU4
& ROUGE & BLEU1 & BLEU4
& ROUGE & BLEU1 & BLEU4
& ROUGE & BLEU1 & BLEU4 \\
\cline{1-16}

\gc{PGG-SLT~\cite{guo2025bridging}}
& \gc{} & \gc{\checkmark} & \gc{\checkmark}
& \gc{52.01} & \gc{53.16} & \gc{27.09}
& \gc{51.85} & \gc{53.45} & \gc{26.85}
& \gc{-} & \gc{-} & \gc{-}
& \gc{-} & \gc{-} & \gc{-} \\ 

\gc{GFSLT-VLP~\cite{yin2023gloss}}
& \gc{} & \gc{\checkmark} & \gc{\checkmark}
& \gc{43.72} & \gc{44.08} & \gc{22.12}
& \gc{42.49} & \gc{43.71} & \gc{21.44}
& \gc{36.44} & \gc{39.20} & \gc{11.07}
& \gc{36.70} & \gc{39.37} & \gc{11.00} \\

\gc{SignCL~\cite{ye2024improving}}
& \gc{} & \gc{\checkmark} & \gc{\checkmark}
& \gc{} & \gc{} & \gc{}
& \gc{49.04} & \gc{49.76} & \gc{22.74}
& \gc{-} & \gc{-} & \gc{-}
& \gc{48.92} & \gc{47.47} & \gc{16.16} \\

\gc{Sign2GPT~\cite{wong2024sign2gpt}}
& \gc{} & \gc{\checkmark} & \gc{\checkmark}
& \gc{-} & \gc{-} & \gc{}
& \gc{48.90} & \gc{49.54} & \gc{22.52}
& \gc{-} & \gc{-} & \gc{-}
& \gc{42.36} & \gc{41.75} & \gc{15.40} \\ 

\gc{GFSLT-VLP-SignCL~\cite{ye2024improving}}
& \gc{} & \gc{\checkmark} & \gc{\checkmark}
& \gc{-} & \gc{-} & \gc{-}
& \gc{49.04} & \gc{49.76} & \gc{22.74}
& \gc{-} & \gc{-} & \gc{}
& \gc{48.92} & \gc{47.47} & \gc{16.16} \\

\gc{SignLLM~\cite{gong2024llms}}
& \gc{} & \gc{\checkmark} & \gc{\checkmark}
& \gc{44.49} & \gc{46.88} & \gc{25.25}
& \gc{47.23} & \gc{45.21} & \gc{23.40}
& \gc{39.18} & \gc{42.45} & \gc{12.23}
& \gc{39.91} & \gc{39.55} & \gc{15.75} \\ 

\gc{LLaVA-SLT~\cite{liang2024llava}}
& \gc{} & \gc{\checkmark} & \gc{\checkmark}
& \gc{-} & \gc{-} & \gc{-}
& \gc{50.44} & \gc{51.20} & \gc{23.43}
& \gc{-} & \gc{-} & \gc{-}
& \gc{51.26} & \gc{52.15} & \gc{20.42} \\ 

\gc{FLa-LLM~\cite{chen2024factorized}}
& \gc{} & \gc{\checkmark} & \gc{\checkmark}
& \gc{-} & \gc{-} & \gc{-}
& \gc{45.27} & \gc{46.29} & \gc{23.09}
& \gc{-} & \gc{-} & \gc{-}
& \gc{37.25} & \gc{37.13} & \gc{14.20} \\

\gc{C2RL~\cite{chen2025c}}
& \gc{} & \gc{\checkmark} & \gc{\checkmark}
& \gc{-} & \gc{-} & \gc{-}
& \gc{50.96} & \gc{52.81} & \gc{26.75}
& \gc{-} & \gc{-} & \gc{-}
& \gc{48.21} & \gc{49.32} & \gc{21.61} \\

\gc{MixSignGraph~\cite{gan2025mixsigngraph}}
& \gc{} & \gc{\checkmark} & \gc{\checkmark}
& \gc{51.71} & \gc{51.07} & \gc{24.87}
& \gc{51.14} & \gc{50.01} & \gc{24.02}
& \gc{49.16} & \gc{49.98} & \gc{20.43}
& \gc{49.93} & \gc{50.24} & \gc{20.78} \\ 

\gc{TwoStream~\cite{chen2022two}}
& \gc{\checkmark} & \gc{\checkmark} & \gc{\checkmark}
& \gc{54.08} & \gc{54.32} & \gc{28.66}
& \gc{53.48} & \gc{54.90} & \gc{28.95}
& \gc{55.10} & \gc{55.21} & \gc{25.76}
& \gc{55.72} & \gc{55.44} & \gc{25.79} \\

\gc{Uni-Sign~\cite{li2025uni}}
& \gc{\checkmark} & \gc{\checkmark} & \gc{\checkmark$^{*}$}
& \gc{-} & \gc{-} & \gc{-}
& \gc{-} & \gc{-} & \gc{-}
& \gc{56.03} & \gc{55.30} & \gc{26.25}
& \gc{56.51} & \gc{55.08} & \gc{26.36} \\

\midrule
\rowcolor[gray]{.85}
\multicolumn{16}{c}{Pose-based} \\
\midrule

VAP~\cite{jiao2024visual}
& \checkmark & & \checkmark
& \underline{51.47} & 52.78 & 26.62
& 51.28 & 53.07 & 26.02
& 51.19 & 53.31 & 23.84
& 51.09 & 52.98 & 23.65 \\

Uni-Sign~\cite{li2025uni}
& \checkmark & & \checkmark$^{*}$
& {48.14*} & {49.14*} & {22.47*}
& {48.10*} & {49.26*} & {22.78*}
& \underline{54.34} & \underline{53.24} & \underline{25.27}
& \underline{53.86} & \underline{54.92} & \underline{25.61} \\

OneSign$^\heartsuit$
& \checkmark & & \checkmark
& \baseline{51.28} & \baseline{52.68} & \baseline{26.38}
& \baseline{51.18} & \baseline{52.68} & \baseline{26.22}
& \baseline{50.05} & \baseline{52.43} & \baseline{24.85}
& \baseline{51.39} & \baseline{51.02} & \baseline{24.42} \\ 

OneSign$^\spadesuit$
& \checkmark & &
& \baseline{52.34} & \baseline{53.41} & \baseline{26.98}
& \baseline{52.01} & \baseline{53.36} & \baseline{26.45}
& \baseline{54.31} & \baseline{53.45} & \baseline{25.28}
& \baseline{53.39} & \baseline{54.45} & \baseline{26.15} \\

\bottomrule
\end{tabular}}  
\end{table*}

\begin{table*}[!t] 
\centering 
\caption{Comparison of gloss-free SLT performance on How2Sign and CSL-OpenWorld.}
\label{tab:GFSLT}   
\resizebox{0.97\textwidth}{!}{
\begin{tabular}{l|lll|lll|lll|lll|lll} 
\toprule
\multirow{3}{*}{Gloss-free SLT}
& \multicolumn{3}{c|}{Extra}
& \multicolumn{6}{c|}{How2Sign}
& \multicolumn{6}{c}{CSL-OpenWorld} \\
& & & 
& \multicolumn{3}{c|}{DEV}
& \multicolumn{3}{c|}{TEST}
& \multicolumn{3}{c|}{DEV}
& \multicolumn{3}{c}{TEST} \\ 
& Pose & RGB & FT
& ROUGE & BLEU1 & BLEU4
& ROUGE & BLEU1 & BLEU4
& ROUGE & BLEU1 & BLEU4
& ROUGE & BLEU1 & BLEU4 \\
\cline{1-16}  

\gc{PGG-SLT (mBART)~\cite{guo2025bridging}}
& \gc{\checkmark} & \gc{\checkmark} & \gc{\checkmark}
& \gc{32.6} & \gc{41.8} & \gc{15.9}
& \gc{31.50} & \gc{38.90} & \gc{13.10}
& \gc{-} & \gc{-} & \gc{-}
& \gc{-} & \gc{-} & \gc{-} \\

\gc{YouTube-SLT$^{*}$~\cite{uthus2023youtube}}
& \gc{} & \gc{\checkmark} & \gc{\checkmark}
& \gc{-} & \gc{-} & \gc{-}
& \gc{-} & \gc{14.96} & \gc{1.22}
& \gc{-} & \gc{-} & \gc{-}
& \gc{-} & \gc{-} & \gc{-} \\ 

\gc{$C^2$RL~\cite{chen2025c}}
& \gc{} & \gc{\checkmark} & \gc{\checkmark}
& \gc{-} & \gc{-} & \gc{-}
& \gc{27.02} & \gc{29.07} & \gc{9.37}
& \gc{-} & \gc{-} & \gc{-}
& \gc{-} & \gc{-} & \gc{-} \\

\gc{FLa-LLM~\cite{chen2024factorized}}
& \gc{} & \gc{\checkmark} & \gc{\checkmark}
& \gc{-} & \gc{-} & \gc{-}
& \gc{27.81} & \gc{29.81} & \gc{9.66}
& \gc{-} & \gc{-} & \gc{-}
& \gc{-} & \gc{-} & \gc{-} \\

\gc{GloFE-VN~\cite{lin2023gloss}}
& \gc{} & \gc{\checkmark} & \gc{\checkmark}
& \gc{12.98} & \gc{15.21} & \gc{2.37}
& \gc{12.61} & \gc{14.94} & \gc{2.24}
& \gc{15.92$^{*}$} & \gc{16.33$^{*}$} & \gc{4.36$^{*}$}
& \gc{15.86$^{*}$} & \gc{16.23$^{*}$} & \gc{4.30$^{*}$} \\

\gc{SSVP-SLT~\cite{rusttowards}}
& \gc{} & \gc{\checkmark} & \gc{\checkmark}
& \gc{-} & \gc{-} & \gc{-}
& \gc{25.70} & \gc{30.20} & \gc{7.00}
& \gc{-} & \gc{-} & \gc{-}
& \gc{-} & \gc{-} & \gc{-} \\

\gc{SignMusketeers~\cite{gueuwou2025signmusketeers}}
& \gc{} & \gc{\checkmark} & \gc{\checkmark}
& \gc{-} & \gc{-} & \gc{-}
& \gc{-} & \gc{18.80} & \gc{4.20}
& \gc{-} & \gc{-} & \gc{-}
& \gc{-} & \gc{-} & \gc{-} \\

\gc{SLT-IV~\cite{tarres2023sign}}
& \gc{} & \gc{\checkmark} & \gc{\checkmark}
& \gc{} & \gc{} & \gc{}
& \gc{-} & \gc{34.01} & \gc{8.03}
& \gc{-} & \gc{-} & \gc{-}
& \gc{-} & \gc{-} & \gc{-} \\ 

\gc{MixSignGraph~\cite{gan2025mixsigngraph}}
& \gc{} & \gc{\checkmark} & \gc{\checkmark}
& \gc{29.24} & \gc{34.82} & \gc{11.28}
& \gc{28.01} & \gc{34.74} & \gc{10.41}
& \gc{22.30$^{*}$} & \gc{23.75$^{*}$} & \gc{8.75$^{*}$}
& \gc{22.32$^{*}$} & \gc{24.00$^{*}$} & \gc{8.91$^{*}$} \\

\gc{SHuBERT~\cite{gueuwouetal2025shubert}}
& \gc{} & \gc{\checkmark} & \gc{\checkmark$^{*}$}
& \gc{-} & \gc{-} & \gc{-}
& \gc{-} & \gc{-} & \gc{16.20}
& \gc{-} & \gc{-} & \gc{-}
& \gc{-} & \gc{-} & \gc{-} \\

\midrule
\rowcolor[gray]{.85}
\multicolumn{16}{c}{Pose-based} \\
\midrule

VAP~\cite{jiao2024visual}
& \checkmark & \checkmark & \checkmark
& 30.27 & 42.34 & 14.91
& 27.77 & 39.22 & 12.87
& - & - & -
& - & - & - \\ 

Uni-Sign~\cite{li2025uni}
& \checkmark & & \checkmark$^{*}$
& - & - & -
& \underline{34.30} & \underline{40.40} & 14.50
& 30.45$^{*}$ & 31.44$^{*}$ & 12.47$^{*}$
& 31.48$^{*}$ & 32.17$^{*}$ & 12.32$^{*}$ \\

OneSign$^\heartsuit$
& \checkmark & &
& \baseline{35.19} & \baseline{39.46} & \baseline{14.92}
& \baseline{36.46} & \baseline{39.47} & \baseline{14.51}
& \baseline{31.65} & \baseline{34.19} & \baseline{13.88}
& \baseline{31.57} & \baseline{33.60} & \baseline{13.48} \\ 

OneSign$^\spadesuit$
& \checkmark & &
& \baseline{37.16} & \baseline{41.56} & \baseline{15.41}
& \baseline{39.15} & \baseline{42.78} & \baseline{15.07}
& \baseline{32.90} & \baseline{35.18} & \baseline{14.71}
& \baseline{32.99} & \baseline{34.83} & \baseline{14.44} \\ 

\bottomrule
\end{tabular}
}   
\end{table*}

%% file: Section/conclusion.tex
In this paper, we present a unified framework that integrates ISLR, CSLR and SLT within a single model. By reformulating all tasks under a unified paradigm, our approach enables joint training and inference across different SL tasks using a single model checkpoint, without task-specific  fine-tuning. To support effective learning from heterogeneous datasets, we introduce the MHCL that reduces vocabulary conflicts while allowing the shared backbone to be jointly optimized. Furthermore, to accommodate the diverse characteristics of sign and text modalities within a shared decoder, we propose a MA-MoE architecture with both shared and modality-specific experts, dynamically activated by a modality-aware router. Extensive experiments on seven public benchmarks across ISLR, CSLR, and SLT tasks demonstrate that our unified framework consistently achieves strong performance, validating its effectiveness   across tasks and datasets.

%% file: Section/sup.tex
\input{table/used_dataset}

\section{Impact Statement}
This paper presents a unified modeling framework for multiple sign language understanding tasks, with the goal of advancing the field of machine learning and sign language processing. In addition to methodological contributions, the work includes a dataset constructed from publicly available online resources, following standard research practices.

Potential positive societal impacts include improved accessibility of information and communication for hard-of-hearing communities, as well as facilitating future research in SL understanding. The dataset is derived from publicly accessible content and is used solely for research purposes. No personally identifiable information is intentionally collected, and the work does not involve interaction with human subjects or user-facing deployment.

Overall, we do not foresee any significant negative ethical or societal consequences arising directly from this work beyond those commonly associated with machine learning research using publicly available data.

\section{Limitations and Discussion} 
We outline potential directions for future improvement.
\paragraph{Heterogeneous datasets training.} While OneSign explicitly models modality-level distribution differences, it does not yet account for distributional discrepancies across heterogeneous datasets. When multiple datasets are jointly used for training, the original data distribution of a specific dataset may be diluted, which can adversely affect performance on its corresponding test set. Consequently, how to effectively balance multi-task and multi-dataset aggregation while preserving dataset-specific characteristics  remains an important direction for future research.

\paragraph{Pose estimation.} Moreover, pose-based representations enable significantly more efficient training and inference, making them particularly suitable for large-scale scenarios. Considering limited GPU resources and the trade-off between training efficiency and model performance, we adopt pose data as the primary input modality instead of raw RGB video. Prior SL models, such as UniSign, CoSign, TwoStream, and MSLU, have demonstrated that pose-based inputs remain effective and robust across multiple benchmarks, suggesting that pose estimation errors do not substantially hinder overall performance in SL tasks. Nevertheless, pose estimation methods (DWPose) may still introduce noise, which can potentially affect model performance.

\input{Section/experiment}

\section{More Ablation Study}

\paragraph{Effect of the Dataset and Task descriptor.}

\input{table/descrip}

\input{table/hypramter}

As illustrated in Fig.~\ref{fig:model2}, we introduce dataset and task descriptors in the MA-MoE module to provide auxiliary conditioning during both training and inference. These descriptors are not used to create separate branches, but rather to guide the model toward the corresponding data distributions in a unified manner.

To evaluate their impact, we conduct an ablation study by selectively removing the dataset and/or task descriptors. The results are summarized in Table~\ref{tab:descriptor}. We observe that removing either descriptor individually leads to only marginal performance degradation, indicating that the model remains robust and effective even without explicit conditioning signals.
Nevertheless, incorporating both descriptors yields the best overall performance, suggesting that they provide complementary information that helps the model better capture dataset-specific and task-specific variations. Importantly, these results demonstrate that our framework does not rely on explicit descriptors to maintain unified modeling capability, while still benefiting from them when available.

\paragraph{Hyperparameter Analysis of the Load Balancing Coefficient.}
We further investigate the sensitivity of the load-balancing coefficient $\lambda$ in Eq.~13. As shown in Table~\ref{tab:lamama}, the best performance is achieved when $\lambda=0.5$, yielding the lowest WER (21.8) and the highest BLEU-4 (26.45).
When $\lambda$ is relatively small (e.g., 0.3), the load-balancing constraint is insufficient to effectively prevent expert imbalance, leading to slightly inferior performance. As $\lambda$ increases beyond 0.5, both WER and BLEU-4 gradually deteriorate. This suggests that an excessively strong balancing constraint forces the router to distribute tokens more uniformly across experts, limiting expert specialization and reducing the routing flexibility required for modeling modality-specific patterns. Overall, these results indicate that a moderate load-balancing coefficient provides the best trade-off between balanced expert utilization and expert specialization.

\paragraph{Comparison of different expert designs}

\input{table/MAMOE_vs_FFN}

To validate the effectiveness of the proposed sparse MA-MoE, we investigate whether the performance gains mainly stem from the MoE routing mechanism or simply from increased parameter capacity. Specifically, we compare MA-MoE with two alternative settings: 
(1) \textbf{Shared + modality-specific FFN}, which consists of a shared expert along with one sign expert and one text expert, but without sparse routing; 
(2) \textbf{Dense FFN}, where three experts are always activated without modality-specific specialization. (3) \textbf{Sparse MoE}, which consists of five experts with Top-$k=3$ activated for each token, but without modality-specific specialization.

As shown in Table~\ref{tab:ComparedMAMOE}, under comparable parameter scales, the Dense FFN achieves the worst performance. This indicates that merely increasing the number of parameters does not lead to performance improvement. In contrast, the proposed sparse modality-specific experts effectively capture modality heterogeneity, demonstrating that the performance gains primarily come from the adaptive routing mechanism rather than parameter scaling.

\input{table/differentLLM}

\paragraph{Effect of MA-MoE on different LLM models.} We further evaluate the generalization capability of MA-MoE across different backbone LLMs. Specifically, we conduct experiments on several representative open-source models, including LLaMA3-1B (our baseline), Gemma-3-1B, and Qwen-2.5-1.5B. As shown in Table~\ref{tab:DiifernetLLM}, MA-MoE consistently improves performance across all evaluated models.  These results demonstrate that our approach is model-agnostic and can be seamlessly integrated into a variety of LLM architectures. Moreover, the consistent gains across different backbones indicate that MA-MoE effectively addresses modality heterogeneity in a generalizable manner, rather than relying on model-specific inductive biases, leading to stable and significant improvements.

\input{table/Expert2_full}

\paragraph{Full ablation on the Impact of Activated Expert Count.}
Due to space limitations in the main paper, we only reported the ablation results on the DEV set in Table~\ref{tab:Expert2} of main paper. Here, we provide the complete results on both DEV and TEST sets in Table~\ref{tab:Expert2_full}.

\section{Qualitative Analysis}
\input{table/qualtitative}

As shown in Table~\ref{tab:CSLR_Qualitative} and Table~\ref{tab:SLTQualitative}, We provide qualitative analyses for both recognition and translation tasks. Specifically, we present recognition results on PHOENIX-2014T and CSL-Daily, and translation results on PHOENIX-2014T, How2Sign, CSL-Daily, and CSL-OpenWorld.
As observed, the model without MA-MoE already achieves reasonable recognition and translation performance. With MA-MoE, OneSign consistently produces more accurate and semantically coherent outputs across datasets, further demonstrating the effectiveness of our approach.

\section{Dataset Visualization}
\begin{figure*}
    \centering
    \includegraphics[width=0.9\linewidth]{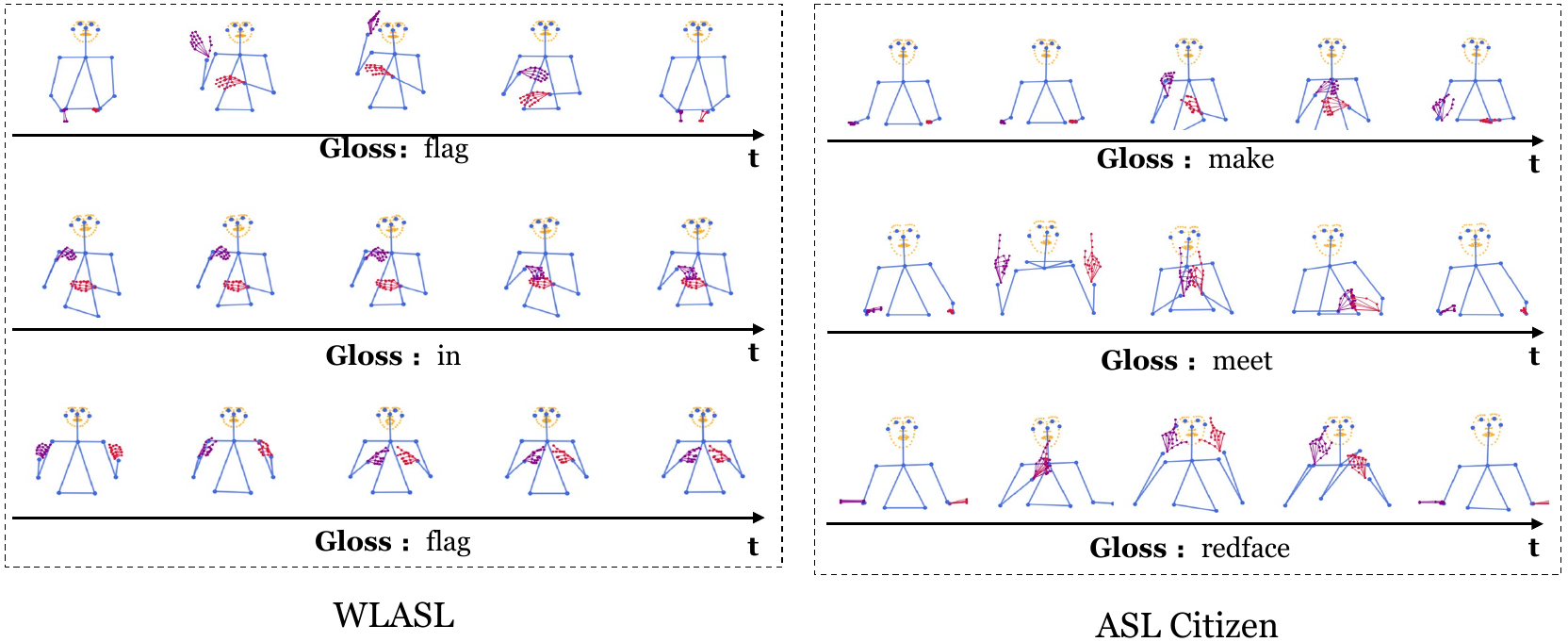}
    \vspace{10mm}
     \includegraphics[width=0.9\linewidth]{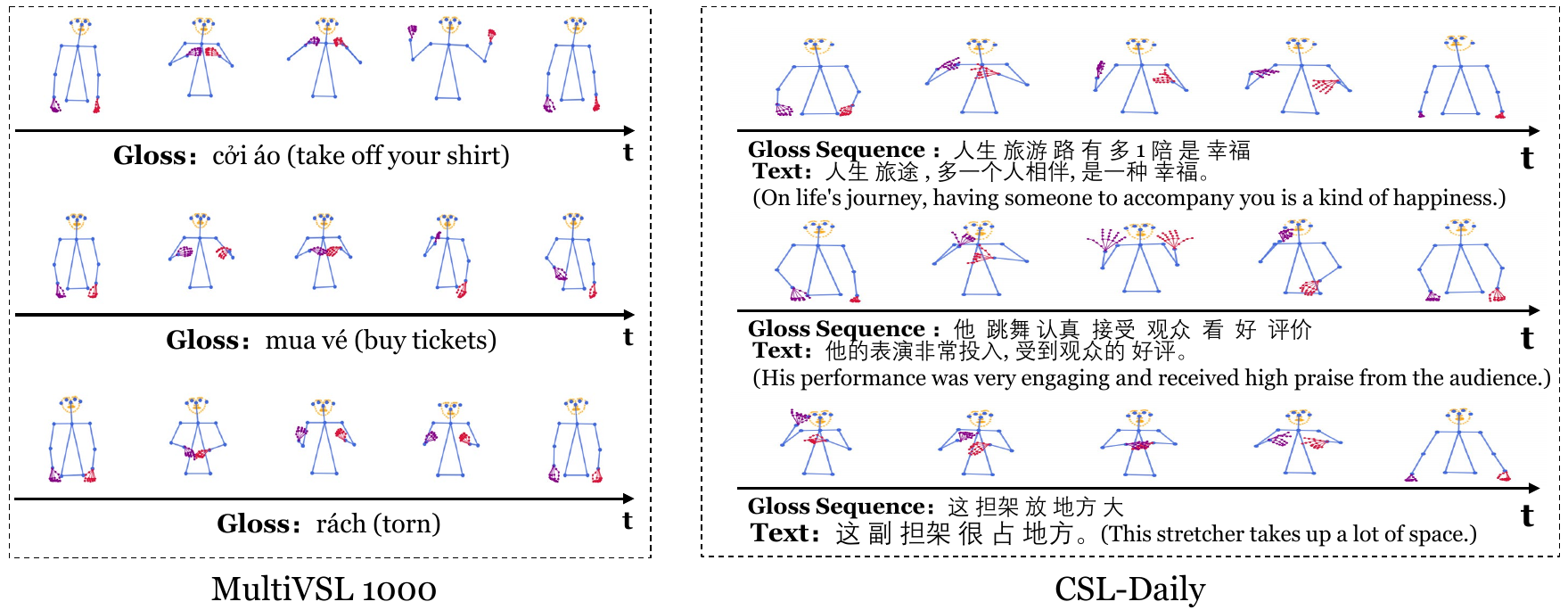}
    \caption{ { Visualization } of used datasets.}
    \label{fig:vis_dataset1}
\end{figure*}

\begin{figure*}
    \centering 

     \includegraphics[width=0.9\linewidth]{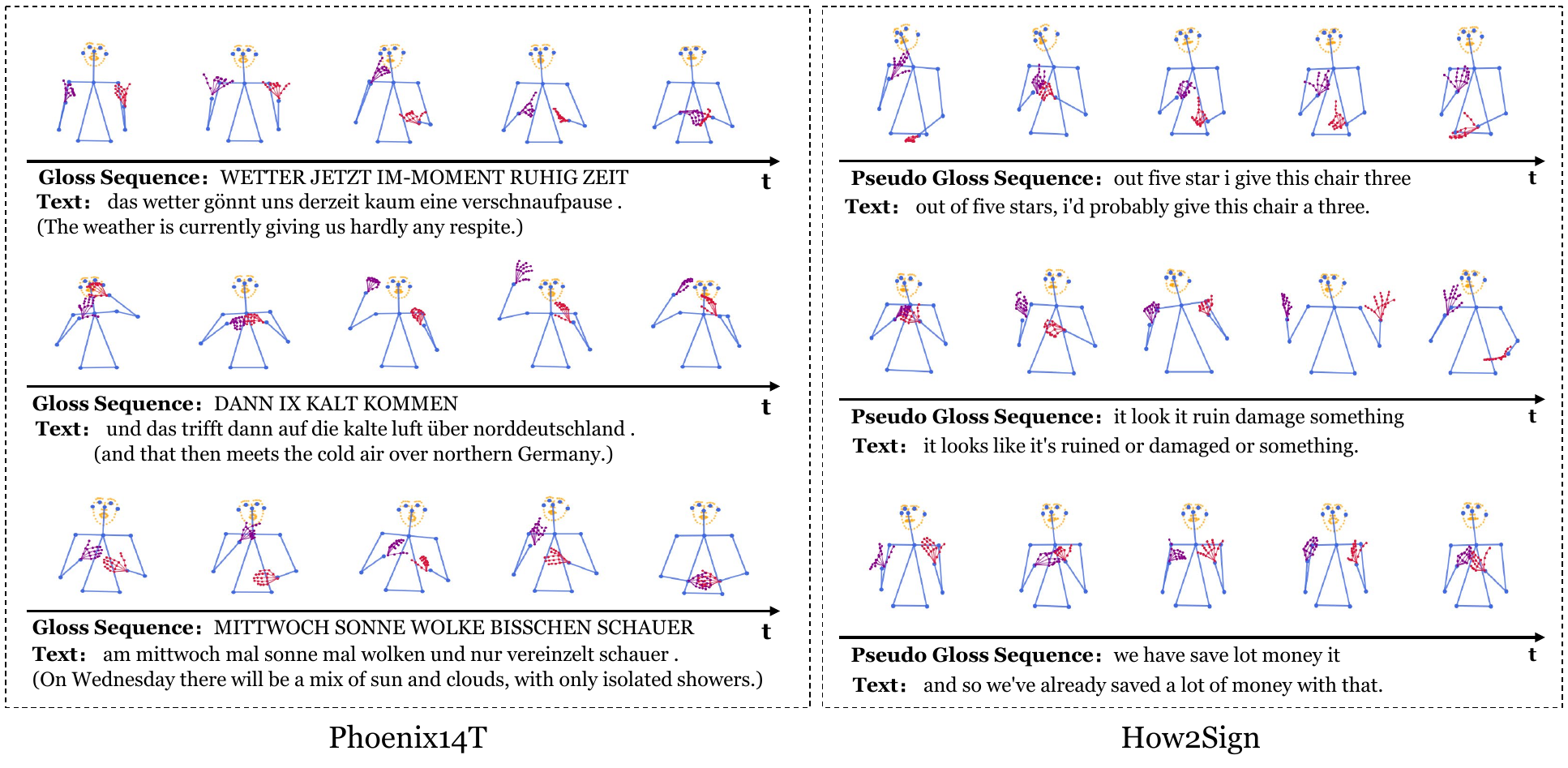}  
    \vspace{3mm}
     \includegraphics[width=0.9\linewidth]{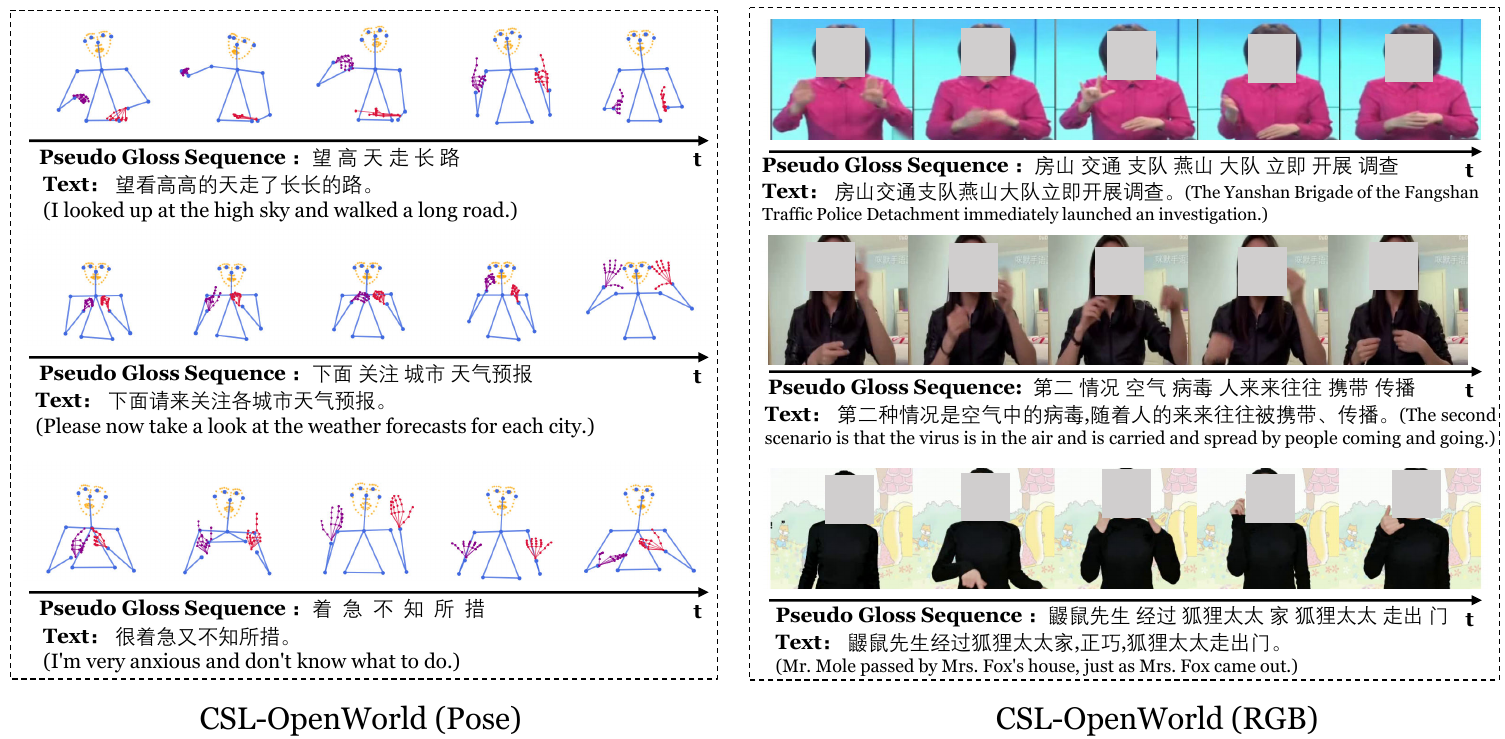}
    \caption{{ Visualization } of used datasets.  We blurred the facial areas for privacy.}
    \label{fig:vis_dataset2}
\end{figure*}

\begin{figure*}
    \centering  
     \includegraphics[width=0.9\linewidth]{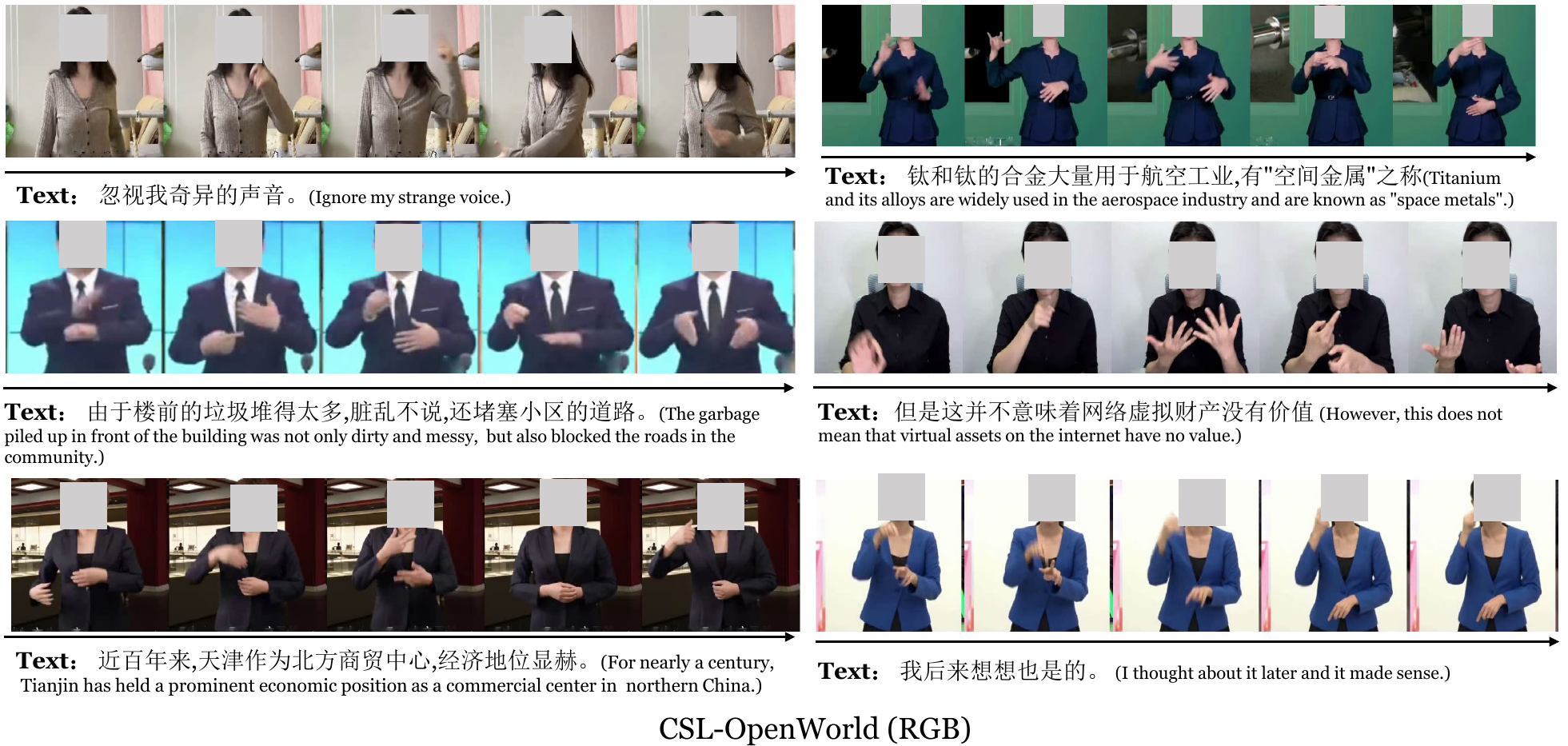} 
       \includegraphics[width=0.9\linewidth]{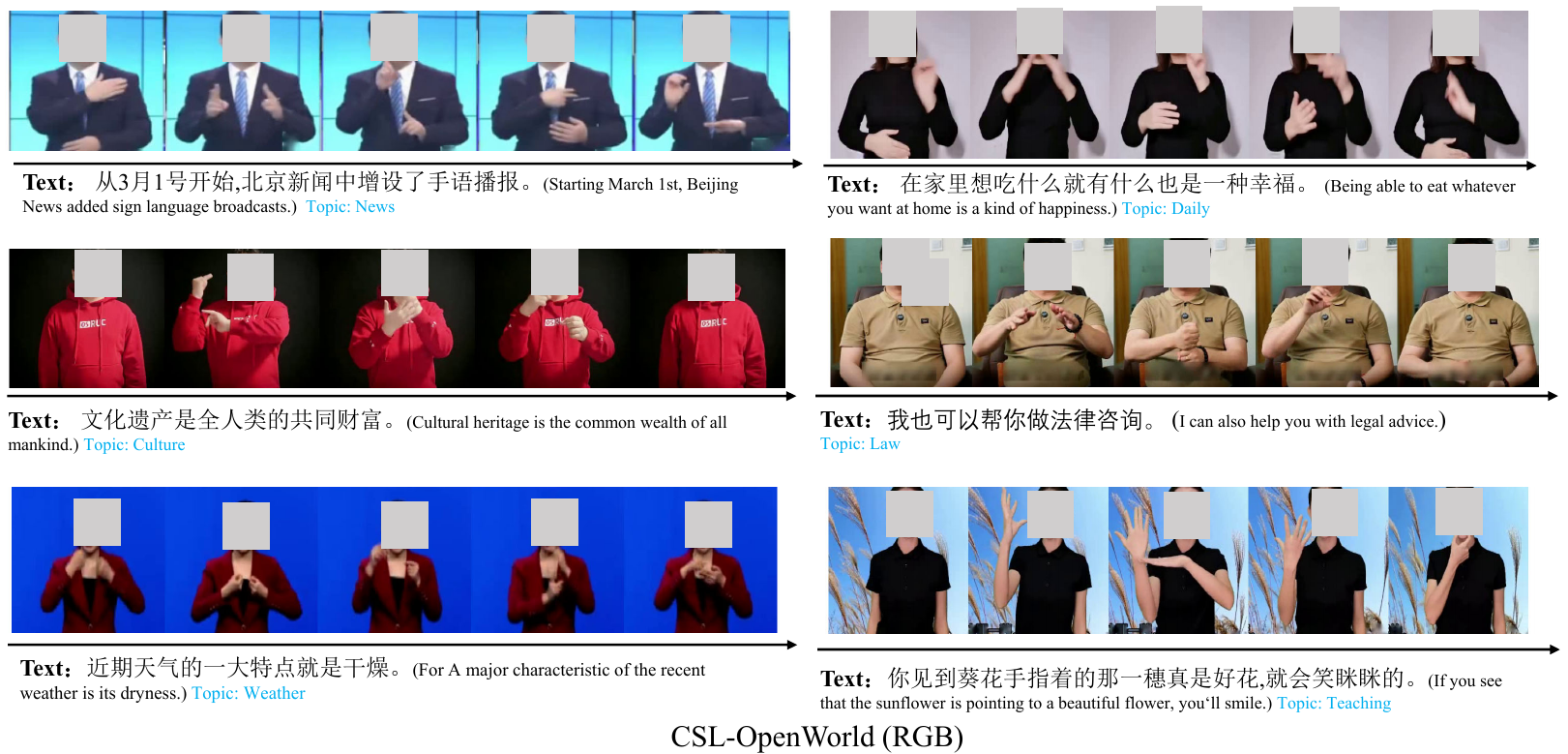}  
\caption{More visualizations of the proposed CSL-OpenWorld dataset, illustrating its diversity in backgrounds, signers, and topics. We blurred the facial areas for privacy.}
    \label{fig:vis_dataset3}
\end{figure*}

We provide visualizations of the formatted pose data used in our experiments across seven datasets. As shown in Figures~\ref{fig:vis_dataset1} and~\ref{fig:vis_dataset2}, the pose representations exhibit noticeable variations in skeleton structure, motion patterns, and temporal dynamics across different datasets. In addition, we present representative video samples from the proposed CSL-OpenWorld dataset, as illustrated in Figure~\ref{fig:vis_dataset3}, to provide an intuitive understanding of its visual diversity and real-world complexity.

%% file: table/used_dataset.tex
\begin{table*}[t]
	\centering 
    	\caption{Details of datasets used in our paper.}  
        \vspace{-3mm}
	\resizebox{0.91\textwidth}{!}{
\begin{tabular}{l|ccc|ccc|ccc} 
	\bottomrule
\multirow{2}*{Dataset}&\multicolumn{3}{c|}{Video Samples}& \multicolumn{3}{c|}{Gloss Vocabulary}&\multicolumn{3}{c}{Token Vocabulary}  \\ 
\cline{2-10}
& {Train}&{Test}&{Validation}& {Train}&{Test}&{Validation}& {Train}&{Test}&{Validation}  \\ 
\toprule 
ASL Citizen~\cite{desai2023asl}  &40154&32941&10304& 2731&2731&2731& -&-&-\\
WLASL2000~\cite{li2020word}  &14289&2878&6149& 2000&1999&2000& -&-&-\\
MultiVSL1000~\cite{dinh2025sign}  &20161&4538&3712& 999&999&999& -&-&-\\
PHOENIX14T~\cite{cihan2018neural}   &7096&642&519& 1085&411&393& 2143&976&927\\
CSL-Daily~\cite{zhou2021improving} &18401&1176&1077&  2000&1345&1358  &2342&1358&1344 \\ 
How2Sign~\cite{duarte2021how2sign}  &30904&2322&1713& -&-&-&8816&3312& 3046 \\ 
CSL-OpenWorld (Ours) &86387&7038&7033&-&-&-&12382&9134&9153\\
	\bottomrule
	\end{tabular}} 
	\label{tab:dataset2}
\end{table*}

%% file: Section/experiment.tex
\section{Experiment}
\label{sec:Imp}

\paragraph{Dataset.}
We adopt seven publicly available datasets, including three ISLR datasets (ASL Citizen, WLASL2000, and MultiVSL1000), two CSLR datasets, which also serve as gloss-based SLT datasets (CSL-Daily and PHOENIX14T), and two gloss-free SLT datasets (How2Sign and our newly proposed CSL-OpenWorld).  Table~\ref{tab:dataset2} summarizes detailed statistics of all datasets used in this paper, including training set size and vocabulary size.

\begin{itemize} 
\item \textbf{WLASL2000}~\cite{li2020word} is the largest public Word-Level ASL dataset. It contains 21,083 video clips performing 2,000 common ASL glosses, featuring 119 signers to ensure variety in signing styles and backgrounds.

\item \textbf{ASL Citizen}~\cite{desai2023asl} is a large-scale, crowdsourced dataset designed for isolated sign language recognition in real-world settings. It comprises approximately 84,000 video clips of 2,731 signs performed by 52 signers, capturing significant diversity in lighting, camera angles, and backgrounds.

\item \textbf{Multi-VSL1000}~\cite{dinh2025sign}  is a multi-view dataset for isolated Vietnamese Sign Language recognition, covering 1,000 sign categories. It provides synchronized observations from multiple viewpoints, enabling research on robust sign recognition under varying viewing conditions.
    
\item \textbf{PHOENIX14T}~\cite{cihan2018neural} is a German Sign Language (DGS) dataset annotated with both glosses and German translations. It consists of 7,096 training, 519 validation, and 642 test samples from 9 signers, with a vocabulary of 1,066 glosses and 2,877 German words.
 
\item \textbf{CSL-Daily}~\cite{zhou2021improving} is a Chinese Sign Language dataset focusing on daily life topics. It contains 18,401 labeled videos for training from 10 signers, providing 2,000 gloss tokens and 2,343 Chinese words for translation tasks.
    
\item \textbf{How2Sign}~\cite{duarte2021how2sign} is a large-scale American Sign Language (ASL) dataset comprising over 80 hours of multi-view video and multimodal data. We utilize the RGB frontal-view videos, which include 31,128 training, 1,741 validation, and 2,322 test samples.  

\item \textbf{CSL-OpenWorld} is our proposed large-scale CSL dataset, collected from diverse online platforms. It includes 86,386 training, 7,033 validation, and 7,038 test samples, featuring over 10K video instances that reflect complex real-world signing scenarios. 

\end{itemize} 

\textit{About dataset, although MSASL is a widely used benchmark for ISLR, we were unable to include it in our experiments. During data acquisition, we found that a substantial portion of the YouTube links provided by MSASL were no longer valid or accessible, making it infeasible to reconstruct the dataset reliably. 
OpenASL contains samples with multiple signers in a single video, which makes it unsuitable for our pose-based modeling. Further processing is required; we plan to incorporate OpenASL results in future work.
Therefore, MSASL and OpenASL is excluded from our experimental evaluation.}

\textbf{Although our experiments involve multiple sign language datasets, they are independently collected and constructed from distinct data sources. We confirm that there is no sample- or video-level overlap across the datasets used in our experiments. Consequently, joint training on these datasets does not introduce cross-dataset leakage through duplicated samples or videos.}

\paragraph{Implementation Details.}
Our architecture consists of following key components.

\noindent\textbf{(1) \textit{Pose Extraction:}}
We employ the open-source DWPose toolkit to extract pose features from videos, including 21 keypoints for each hand, 68 facial keypoints, and 14 body joint keypoints. 

\noindent\textbf{(2) \textit{MA-LLM MoE:}}
We adopt LLaMA~3.2 (1B) as the backbone of our modality-adaptive LLM with a Mixture-of-Experts (MoE) design. As the default setting, we use two pose experts with one activated expert and two text experts with one activated expert. The dataset descriptor indicates the sign language video dataset, while the task descriptor specifies the corresponding task (\ie Isolated Sign Language Recognition, Continuous Sign Language Recognition, and Sign Language Translation).

\noindent\textbf{(3) \textit{Unified Paradigm:}}
Following prior work~\cite{gan2025mixsigngraph}, pseudo-gloss annotations are constructed by removing punctuation, performing lemmatization, and applying word-level tokenization to the text transcripts. In all experiments, the loss weights $\lambda_{\text{sign}}$ and $\lambda_{\text{text}}$ are both set to 0.5.

\noindent\textbf{(4) \textit{Optimization:}}
Our model is implemented in PyTorch~2.6 and trained on four NVIDIA RTX~4090 GPUs (24GB memory each) using half-precision training. The initial learning rate is set to $1\times10^{-4}$ for the recognition module and $1\times10^{-5}$ for the translation module, with a decay factor of 0.5 applied at epochs 20, 30, and 40. The batch size is set to 8.
Notably, the model is trained without incorporating any additional external sign language datasets beyond those described above.

\paragraph{Evaluation Metrics.}
For ISLR, to ensure fair comparison with prior work and their reported baselines, we follow the evaluation protocols used in each dataset. Specifically, we report Top-1 and Top-5 accuracy (Top@1 and Top@5) on ASL Citizen and MultiVSL1000, and per-instance (P-I) and per-class (P-C) Top-1 accuracy on WLASL2000. 
For CSLR, we use Word Error Rate (WER) as the evaluation metric. 
For SLT, we adopt ROUGE-L F1 and BLEU-1/2/3/4 to evaluate translation quality. 
These metrics are widely used in existing CSLR and SLT literature, enabling direct and fair comparison with prior methods.

%% file: table/descrip.tex
\begin{table}[!t] 
	\centering 
   \captionof{table}{\label{tab:descriptor}Effect of Dataset and Task
descriptor.} 
\vspace{-2mm}
\resizebox{0.96\columnwidth}{!}{ 
		\begin{tabular}{cc|cccc|cccc}
		\shline
\multirow{2}*{Dataset}&\multirow{2}*{Task} &\multicolumn{4}{c|}{Dev} &  \multicolumn{4}{c}{TEST} \\
&&WER&Rouge& BLEU1& BLEU4&WER&Rouge& BLEU1& BLEU4 \\
\shline  
  & &20.1&51.58&52.12& 26.01&20.2&51.59	&52.01	&25.64\\
\checkmark  &&20.0&52.24&53.14& 26.76&20.2&51.87	&53.16	&26.31 \\
 &\checkmark& 20.0&52.13&53.35& 26.65&20.1&51.64	&53.07	&26.14\\
\checkmark &\checkmark &\baseline{20.0}& \baseline{52.34}&\baseline{53.41}& \baseline{26.98}&\baseline{20.1}&  \baseline{52.01}&\baseline{53.36}&\baseline{26.45}\\	 	
	\shline 
	\end{tabular} }
\end{table}

%% file: table/hypramter.tex
\begin{table}
\caption{ Phoenix14T results with $\lambda$ in $\mathcal{L}_{\mathrm{bal}}$ .}
\centering  
\resizebox{0.5\linewidth}{!}{
\begin{tabular}{l|lllll}
\hline
$\lambda_{sign}=\lambda_{text}$ & 0.3 &0.5&0.7&0.9&1\\
\hline 
WER$\downarrow$ &22.1&20.1 &22.3 &23.4&25.1 \\
BLUE4 $\uparrow$&25.75 &26.45 &25.74 &25.15&24.78 \\
\hline
\end{tabular}}
\label{tab:lamama}  
\end{table}

%% file: table/MAMOE_vs_FFN.tex
\begin{table*}[!t] 
	\centering 
   \captionof{table}{\label{tab:ComparedMAMOE}Comparison of different experts.} 
      \vspace{-2mm}
\resizebox{0.9\linewidth}{!}{ 
		\begin{tabular}{c|cc|cccc|cccc}
		\shline
\multirow{2}*{Model}&\multicolumn{2}{c|}{Parameter(M)}&\multicolumn{4}{c|}{Dev}&\multicolumn{4}{c}{TEST} \\
&Total&Activated&WER&Rouge& BLEU1& BLEU4&WER&Rouge& BLEU1& BLEU4 \\
\shline  
Dense FFN &2041.1&2041.1&22.6 &49.63&50.23  &23.68 &22.8 &49.56&	49.32&23.67\\	
Sparse MOE &4457.0&2041.1&{22.4}& {49.23}&{49.36}& {23.15}&{22.6}&  {48.65}&{49.36}&{22.64}\\	
shared+modality-specific FFN &2041.1&2041.1&21.4  &51.69 & 51.60& 25.68 &21.5 & 51.52&51.98&25.79\\
MA-MOE &4457.0&2041.1&\baseline{20.0}& \baseline{52.34}&\baseline{53.41}& \baseline{26.98}&\baseline{20.1}&  \baseline{52.01}&\baseline{53.36}&\baseline{26.45}\\	 		
	\shline 
	\end{tabular} }
\end{table*}

%% file: table/differentLLM.tex
\begin{table*}[!t] 
	\centering 
   \captionof{table}{\label{tab:DiifernetLLM}Effect of MA-MOE on different LLMs.}
      \vspace{-2mm}
\resizebox{0.9\linewidth}{!}{ 
		\begin{tabular}{c|cccc|c|cccc}
		\shline
\multirow{2}*{Model}&\multicolumn{4}{c|}{Test}&\multirow{2}*{+MA-MOE}&\multicolumn{4}{c}{TEST} \\
&WER&Rouge& BLEU1& BLEU4& &WER&Rouge& BLEU1& BLEU4 \\
\shline  
Llama 3.2 1B &22.5 & 48.99 & 49.72 & 23.01&$\rightarrow$ &{20.1}   &{52.01}&{53.36}&{26.45}\\	
Gemma3 1b    &22.9  &49.96   &48.04   &22.85 &$\rightarrow$ &{20.2}   &51.32   &52.66	  &26.35\\	
Qwen2.5 1.5B &23.7  &49.06   &48.81   &22.93 &$\rightarrow$ &{20.3}   &51.17   &52.69	  &26.34\\
	\shline 
	\end{tabular} }
\end{table*}

%% file: table/Expert2_full.tex
\begin{table*}[!t] 
	\centering  
   \captionof{table}{\label{tab:Expert2_full} {Impact of activated expert count.}} 
   \vspace{-2mm}
        \resizebox{0.9\linewidth}{!}{
\begin{tabular}{cc|rr|ccc|ccc}
		\shline
$N_s=N_t$ & Top-$k$ & Params (M) & Activated (M)& Rouge & { BLEU1 }& { BLEU4 }& Rouge & { BLEU1 }& { BLEU4 } \\
	\shline 
0 & 0 & 1235.8& 1235.8 &49.45&49.36&22.67& {48.99}&{49.72}&{23.01}\\
2 & 1 & 4457.0 & 2041.1	&52.34& 53.41 &26.98 & {52.01}&{53.36}&{26.45} \\
4 & 1 &  7678.2& 2041.1 &52.74&	53.65&	26.78& {52.14}&{53.24}&{26.53}\\
4 & 2 &  7678.2 &  2846.4 &52.99	&53.94	&26.97	&{52.45}&{53.47}&{26.56}\\
	\shline 
\end{tabular}
}
\end{table*}

%% file: table/qualtitative.tex
\begin{CJK}{UTF8}{gbsn} 

\begin{table*}[t]
	\centering  
      \captionof{table}{CSLR  Qualitative results on Phoenix-2014T and CSL-Daily.}
	\resizebox{0.6\textwidth}{!}{
		\centering
		\begin{tabular}{l|l}
		\hline 
           example(a)&Phoenix14T dataset\\
       \hline 
        Groundtruth&HEUTE \quad NACHT\quad REGEN\quad DOCH \quad MITTE \quad NORD\\
        &(TODAY \quad NIGHT \quad RAIN \quad HOWEVER \quad MIDDLE \quad NORTH)\\
        OneSign & { HEUTE \quad NACHT \quad REGEN \quad DOCH \quad MITTE \quad NORD}\\
        & {TODAY \quad NIGHT \quad RAIN \quad HOWEVER \quad MIDDLE \quad NORTH}\\
        W/O MA-MOE &  HEUTE \quad NACHT \quad NEBEL \quad  DOCH \quad MITTE  \quad NORD \\
        & ( TODAY \quad NIGHT \quad  FOG \quad HOWEVER \quad MIDDLE  \quad NORTH )\\
      \hline  
     example(b)&Phoenix14T dataset\\
       \hline 
        Groundtruth&TAG \quad  OST \quad  VIER \quad  FLUSS \quad  SIEBEN  \quad  GRAD\\
        &(DAY \quad  EAST \quad FOUR\quad RIVER \quad SEVEN \quad DEGREE)\\
        OneSign & { TAG \quad  DANN \quad  VIER \quad  FLUSS \quad  SIEBEN  \quad  GRAD  \quad NORD}\\
        & {(DAY \quad THEN \quad FOUR\quad RIVER\quad SEVEN \quad DEGREE \quad NORTH)}\\
        W/O MA-MOE &{ TAG \quad  DANN \quad  VIER \quad  FLUSS \quad  SIEBEN } \\
        & ( DAY \quad THEN \quad FOUR\quad RIVER\quad SEVEN)\\
      \hline  
        example(c)&CSL-daily dataset\\
       \hline 
        Groundtruth&去 \quad学校 \quad我\quad 踢足球\\
        & (Go\quad School \quad I \quad Play soccer)\\
        OneSign& {去 \quad 学校\quad 我\quad 踢足球}\\
         &{Go \quad School \quad I \quad Play  soccer}\\
         W/O MA-MOE &来 \quad 学校 \quad  我 \quad 踢足球 \\
      &Come \quad School \quad I \quad Play  soccer\\
            \hline  
      example(d)&CSL-daily dataset\\
               \hline 
        Groundtruth&我 \quad 房子\quad  在\quad  看 \quad 电视\quad  你\quad  看 \quad 电视 \quad 喜欢\\
        & (I \quad  House \quad  At \quad  Watch  TV \quad  You \quad  Watch \quad TV \quad  Like)\\
        OneSign& {我\quad   房子\quad  看\quad  电视 \quad 你 \quad 看\quad  电视 \quad 喜欢}\\
         &{I \quad House \quad   Watch \quad  TV \quad You\quad  Watch \quad TV \quad Like}\\
         W/O MA-MOE &我\quad  房子 \quad 在\quad  看 \quad 电视\quad  我\quad  看 \quad 电视 \\
      &I \quad House \quad At\quad  Watch \quad  TV\quad  I\quad  Watch \quad TV\\
            \hline  
      \hline
	\end{tabular}}
\label{tab:CSLR_Qualitative}
\end{table*}

\begin{table*}[t]
	\centering  
      \captionof{table}{SLT  Qualitative results on Phoenix-2014T, How2Sign , CSL-Daily and CSL-OpenWorld.}
	\resizebox{0.9\textwidth}{!}{
		\centering
		\begin{tabular}{l|l}
		\hline 
           example(a)&Phoenix14T dataset\\
       \hline 
        Groundtruth&und nun die wettervorhersage für morgen sonntag den zwölften juli\\
        &(And now the weather forecast for tomorrow, Sunday, July 11th)\\
        OneSign & {\text{ und nun die wettervorhersage für morgen sonntag den zwölften juli}}\\
        & {\text{ (And now the weather forecast for tomorrow, Sunday, July 11th)}}\\
        W/O MA-MOE & und nun die wettervorhersage für morgen sonntag den {\color{red}{elften}} juli   \\
        & (And now the weather forecast for tomorrow, Sunday, July 12th)\\
    \hline 
	example(b)&How2Sign dataset\\
     \hline 
       Groundtruth&So, we've got to find a way to get to the exit.\\
       OneSign&{we've got to find a way to get to the exit.}\\  
        W/O MA-MOE & we  have found a way to get to the exit. \\  
      \hline  
        example(b)&CSL-daily dataset\\
       \hline 
        Groundtruth&人生旅途, 多一个人相伴, 是一种幸福。\\
        & (On the journey of life, having someone by your side is a form of happiness.)\\
        OneSign& {人生旅途, 多个人相伴, 是一种幸福。}\\
         &{On the journey of life, having others by your side is a form of happiness.}\\
         W/O MA-MOE &人生旅途, 相伴是一种幸福。 \\
      &On the journey of life, companionship is a blessing.\\
            \hline  
        example(d)& CSL-OpenWorld dataset\\
       \hline 
        Groundtruth&文化遗产是全人类的共同财富。\\
        & (Cultural heritage is the common wealth of all humanity.)\\
        OneSign& {文化遗产是人类的共同财富。}\\
         &{Cultural heritage is the common wealth of humanity.}\\
         W/O MA-MOE &文化是全人类的共同财富。 \\
      &Culture is the common wealth of all humanity.\\
      \hline
	\end{tabular}}
\label{tab:SLTQualitative}
\end{table*} 

\end{CJK}